%% file: UAT_MC.tex
\documentclass{article}
\usepackage{ijcai26}

\usepackage{times}
\usepackage{soul}
\usepackage{url}
\usepackage[hidelinks]{hyperref}
\usepackage[utf8]{inputenc}
\usepackage[small]{caption}
\usepackage{graphicx}
\usepackage{amsmath}
\usepackage{amsthm}
\usepackage{booktabs}
\usepackage{algorithm}
\usepackage{algpseudocode}
\usepackage{amsmath} 

\usepackage[switch]{lineno}
\usepackage{multirow}
\usepackage{subfigure}

\title{Band Together: Untargeted Adversarial Training with \\ Multimodal Coordination against Evasion-based Promotion Attacks}

\author{
Guanmeng Xian$^1$
\and
Ning Yang$^{1}$\thanks{Corresponding author.}
\and
Philip S. Yu$^2$\\
\affiliations
$^1$Sichuan University, Chengdu, China\\
$^2$University of Illinois at Chicago, USA\\
\emails
xianguanmeng@stu.scu.edu.cn,
yangning@scu.edu.cn,
psyu@uic.edu
}

\begin{document}

\maketitle

\begin{abstract}
    Multimodal recommender systems exploit visual and textual signals to alleviate data sparsity, but this also makes them more vulnerable to evasion-based promotion attacks. Existing defenses are largely limited to single-modal settings and mainly focus on poisoning-based threats, leaving evasion-based threats underexplored. 
     In this work, we first identify a cross-modal gradient mismatch under the multi-user promotion setting, where visual and textual perturbations are optimized in inconsistent directions due to the dominance of distinct user groups. This phenomenon dilutes the attack effectiveness and leads robust training to underestimate worst-case risks. To address this issue, we propose \textbf{U}ntargeted \textbf{A}dversarial \textbf{T}raining with \textbf{M}ultimodal \textbf{C}oordination (UAT-MC). UAT-MC tackles the challenge of unknown targeted items in evasion-based attacks (as opposed to poisoning-based attacks) by treating all items as potential targets, and introduces a gradient alignment mechanism to explicitly correct this mismatch. This design ensures synchronized perturbations across modalities, thereby maximizing adversarial strength for robust training. 
    Extensive experiments demonstrate that UAT-MC significantly improves robustness against promotion attacks while maintaining acceptable recommendation performance under the defense–accuracy trade-off. Code is available at \url{https://github.com/gmXian/UAT-MC}.
     
\end{abstract}

\section{Introduction}
 \label{sec:introduction}

\begin{figure}[!t]
\centering
\includegraphics[scale=0.25]{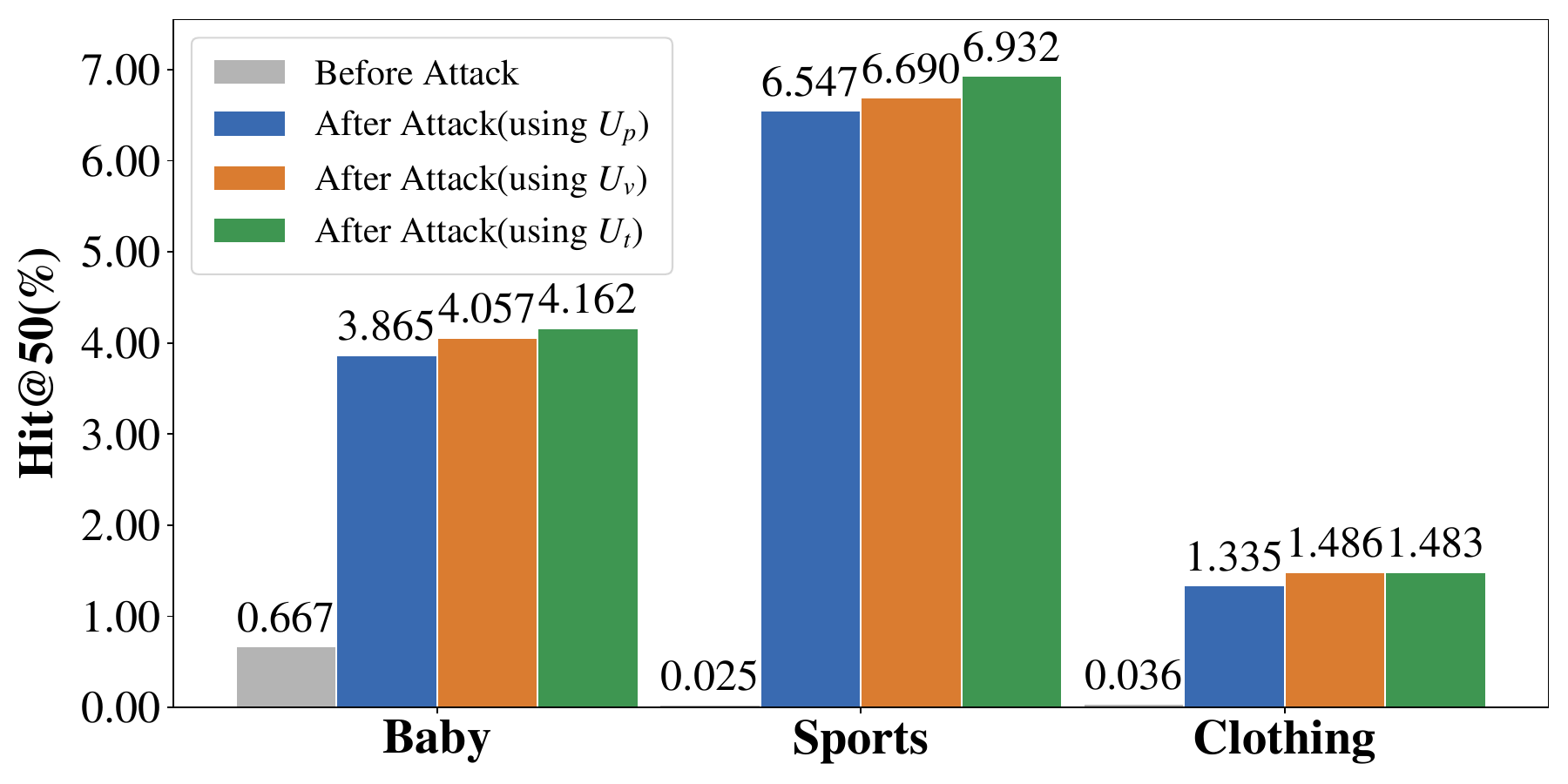} 
\vspace{-1mm}
\caption{VBPR's vulnerability to vanilla FGSM-based promotion attacks. For each dataset, we report the Hit@50 before and after attack under different user subsets, including $\mathcal{U}_p$, $\mathcal{U}_v$, and $\mathcal{U}_t$. 
}
\label{Fig-motivation}
\end{figure}

Multimodal Recommender Systems (MRSs) typically leverage visual and textual information from user-interacted items to capture users' fine-grained preferences, effectively alleviating the challenge posed by sparse interaction data \cite{he_2016_VBPR,wei_2019_MMGCN,zhang_2021_mining_LATTICE,zhou_2023_bootstrap_BM3,Liu_2024_AlignRec}. While auxiliary modal information enhances the personalization of recommendations, we find that MRSs exhibit heightened vulnerability to \textbf{evasion-based promotion attacks}, in which attackers imperceptibly perturb the modalities of targeted items to boost their rankings, as demonstrated by our motivation experiment in Figure \ref{Fig-motivation}. 

Specifically, we conduct evasion-based promotion attacks (FGSM \cite{goodfellow_2014_explaining_FGSM} and PGD \cite{madry2017towards_PGD}) on both visual and textual modalities of unpopular items (defined as those with only 5 interactions) over various datasets (Amazon Baby, Sports and Clothing \cite{he_2016_ups_Amazon}) and architectures (VBPR \cite{he_2016_VBPR} and MMGCN \cite{wei_2019_MMGCN})). Figure \ref{Fig-motivation} presents the results on VBPR under FGSM attack as a representative instance. The black numbers represent the average hit rate ($\text{Hit}@50$) of targeted items appearing in user recommendation lists. As one can find from Figure \ref{Fig-motivation}, the post-attack $\text{Hit}@50$ always shows a significant increase, demonstrating the vulnerability of MRS to evasion-based promotional attacks.

Existing works \cite{wu_2021_Fight_APT,zhang_2024_stealthy_RecTextAttack,mu_2025_trust_Trust-GRS} have attempted to mitigate the vulnerability to promotion attacks; however, they suffer from two critical limitations when applied to MRSs. \textbf{First}, these approaches primarily focus on single-modal recommender systems, failing to address the complex vulnerability patterns that arise from interactions between visual and textual features. This oversight is especially problematic as attackers can exploit cross-modal correlations to amplify their attack impact. For example, the Re-writing Defense \cite{zhang_2024_stealthy_RecTextAttack} uses GPT-3.5-turbo to rewrite adversarial text as a defense for LLM-based recommendation models. \textbf{Second}, and more fundamentally, current defenses are almost exclusively designed for poisoning-based attacks,  where adversaries compromise collaborative signals by injecting fake user profiles or interactions during the training phase. For example, APT \cite{wu_2021_Fight_APT}  simulates the poisoning process by injecting fake user data to foster a more robust system. 
However, in MRSs, attackers can deliberately manipulate the description or image of a targeted item during the inference phase to influence the recommendation results—an aspect not addressed by existing defense methods. To bridge these critical gaps, we propose \textbf{Untargeted Adversarial Training with Multimodal Coordination (UAT-MC)}, a novel adversarial training framework specifically designed to defend against evasion-based promotion attacks in MRSs. However, directly applying conventional adversarial training to this setting is non-trivial due to the following two challenges:

 \textbf{C1: Dynamic Attack Target.} Unlike poisoning attacks where malicious targets are explicitly injected during training, evasion attacks dynamically select targeted items at the inference phase. This fundamental difference renders traditional targeted adversarial training approaches ineffective, as they rely on pre-defined attack targets to generate adversarial examples, which are unknown during the training phase.

 \textbf{C2: Cross-modal Gradient Mismatch.}  Under the multi-user promotion setting, MRSs present a unique challenge: combining perturbations across visual and textual modalities often yields suboptimal adversarial examples which degrade the robustness achieved through adversarial training. Cross-modal gradient mismatch arises when visual and textual perturbations in multi-user promotion attacks are dominated by different user groups, causing the two modalities to be optimized toward inconsistent objectives and resulting in misaligned gradient directions.

To tackle the challenge \textbf{C1}, we propose \textbf{untargeted adversarial training} for MRS, which treats all items as potential targets of evasion-based attacks. Our approach frames recommendation as a multi-label classification task over users, where items serve as labels. Untargeted adversarial training indirectly defends against targeted attacks by globally enhancing the robustness of decision boundaries and disrupting the local gradient information on which attacks rely. The core principle is that the model learns to resist perturbations from any direction, thereby covering attacks originating from specific directions. To tackle challenge \textbf{C2}, we propose a novel \textbf{gradient-aligned multimodal perturbation method} to address cross-modal gradient mismatch in adversarial training. Unlike perturbing visual and textual features independently, which can diminish attack strength due to misaligned gradients—our method enforces gradient synchronization across modalities by minimizing the cosine distance between visual and textual gradients through a joint loss term. This ensures perturbations coherently push items toward adversarial regions. As shown in later experiments, this coordinated approach maximizes attack potency during training and enhances adversarial robustness against evasion-based promotion attacks.

Our contributions can be summarized as follows:
\begin{itemize}

\item We identify evasion-based promotion attacks as a critical threat to multimodal recommender systems, in which adversaries perturb both the visual and textual features of targeted items. 

\item We propose a novel Untargeted Adversarial Training with Multimodal Coordination (UAT-MC) framework to enhance the robustness of MRSs against evasion-based promotion attacks. Specifically, the proposed untargeted adversarial training inherently defends against targeted attacks by universally hardening the decision boundaries, thereby covering all potential attack directions. 

\item We propose gradient-aligned multimodal perturbations to resolve cross-modal gradient mismatch in adversarial training. By minimizing cosine distance between modalities via a joint loss term, our method synchronizes perturbations to maximize attack potency. 

\item Extensive experiments conducted on real datasets verify the effectiveness of our method.

\end{itemize}

\section{Preliminary}
    \label{Preliminary}
In this section, we conceptually define MRS and describe the evasion-based promotion attacks. 

\subsection{MRS}
Let $\mathcal{U}$ and $\mathcal{I}$ denote user set and item set, respectively. $R \in \{0,1\}^{|\mathcal{U}| \times |\mathcal{I}|}$ is user-item interaction matrix, where an element $r_{u,i}$ is 1 if there exists interaction between user $u \in \mathcal{U}$ and item $i \in \mathcal{I}$, otherwise 0. Each item $i$ is associated with a visual modality embedding $\mathbf{v}_i$ and a textual modality embedding $\mathbf{t}_i$ which are generated by CNN and Transformer, respectively. 

Generally, an MRS $f$ infers the probability $\hat{r}_{u,i}$ that user $u$ will interact with item $i$ based on given $R$ and the modality embeddings $\mathbf{v}_i$ and $\mathbf{t}_i$, i.e., 
\begin{equation}
\hat{r}_{u, i}= f(R,u, i, \mathbf{v}_i, \mathbf{t}_i).
\label{eq:MRS}
\end{equation}
Let $\mathcal{D} = \{ (u, i_{+},i_{-}) \}$ be a training set, where $i_{+}$ is $u$'s positive sample ($r_{u,i_{+}}=1$ ) and $i_{-}$ is $u$'s negative sample ($r_{u,i_{-}}=0$ ). Like traditional recommender systems, an MRS is usually trained with Bayesian Personalized Ranking (BPR) \cite{Rendle_2009_BPR} loss function, 
\begin{equation}
\mathcal{L}_\text{BPR}(\Theta) = \sum_{(u,i_{+},i_{-}) \in \mathcal{D}} -\ln \sigma(\hat{r}_{u,i_{+}} - \hat{r}_{u,i_{-}}),
\label{eq:normal_BPR}
\end{equation}
where $ \Theta $ represents the model parameters and $\sigma(\cdot)$ is the sigmoid function.

\subsection{Threat Model}
\subsubsection{Attacker's Objective}  
Given a well-trained MRS, an evasion-based promotion attack is to imperceptibly perturb the multimodal embeddings of a targeted item $i$, so that $i$ will appear in the top-$K$ recommendation lists of as many users as possible. Inspired by \cite{wang_2024_unveiling_CLeaR}, we define the promotion utility function as:
\begin{equation}
	\mathcal{L}_{\text{promotion}} = \frac{1}{N_p}\sum_{u \in \mathcal{U}_{\text{p}}}{
		\text{Sigmoid}(y_{u,i}-y_{u,K}),
	} 
	\label{eq:L_promotion}
\end{equation}
where $y_{u,K}$ denotes the score of the $K$-th ranked item for user $u$. By maximizing Equation (\ref{eq:L_promotion}), the attacker encourages the targeted item score $y_{u,i}$ to surpass the top-$K$ threshold $y_{u,K}$.

\subsubsection{Attacker's Knowledge}  In this paper, we assume a white-box scenario for the generation of the adversarial examples in our UAT-MC, where the attacker has full access to the MRS $f$, including its parameters and gradients. This is because UAT-MC aims to train a more robust MRS, rather than to conduct attacks from the perspective of malicious merchants. This setting represents a worst-case attacker and allows us to evaluate the upper bound of the potential impact of evasion-based promotion attacks.

\subsubsection{Attacker's Capability} Following the ideas of the related works ~\cite{Tang_2020_Adversarial_AMR,zhang_2021_mining_LATTICE,guo_2024_lgmrec_LGMRec,Liu_2024_AlignRec,Ong_2025_Spectrum_SMORE} , we perturb the multimodal embeddings instead of the raw inputs, which allows more fine-grained and efficient manipulations. Formally, the perturbed embeddings $\mathbf{v}'_i = \mathbf{v}_i + \Delta_v^i$ and $\mathbf{t}'_i = \mathbf{t}_i + \Delta_t^i$. For imperceptibility, the perturbations $\Delta_m^i$ ($m \in \{v, t\}$) are constrained by $\|\Delta_m^i\|  \leq \epsilon_m$, where $ \epsilon_m $  is perturbation budget.

\section{Cross-modal Gradient Mismatch Analysis}
\label{sec:gradient_mismatch}

In this section, we demonstrate that under multi-user promotion settings, the gradients of visual and textual perturbations are inherently driven by distinct user groups due to varying modal sensitivities. Consequently, simply aggregating these gradients leads to conflicting optimization directions across modalities,  which we term \textbf{cross-modal gradient mismatch}. This misalignment fundamentally limits the synergy of multimodal attacks and motivates our UAT-MC framework. We next provide a formal analysis to characterize and verify the existence of this mismatch. 

\subsection{Objective Inconsistency Across Modalities}
Given the promotion objective $\mathcal{L}_{\text{promotion}}$ defined over the user set $\mathcal{U}_p$ (Equation~(\ref{eq:L_promotion})), the optimization problem can be formulated as:
\begin{equation}
\max_{\Delta_v^i, \Delta_t^i} \;
\sum_{u \in \mathcal{U}_p} \mathcal{L}_u(\Delta_v^i, \Delta_t^i),
\end{equation}
where $\mathcal{L}_u$ denotes the promotion loss contributed by user $u$.

Accordingly, the perturbations are updated by aggregating gradients from all selected users:
\begin{equation}
G^v = \sum_{u \in \mathcal{U}_p} g_u^v,
\quad
G^t = \sum_{u \in \mathcal{U}_p} g_u^t,
\end{equation}
where
$g_u^v = \frac{\partial \mathcal{L}_u}{\partial \Delta_v^i}$ and
$g_u^t = \frac{\partial \mathcal{L}_u}{\partial \Delta_t^i}$
denote the visual and textual gradients induced by user $u$, respectively.

To quantify how much each user contributes to the final update direction, we define the \textbf{directional contribution} of user $u$ on modality $m \in \{v,t\}$ as:
\begin{equation}
c_u^m = \cos(g_u^m, G^m) \cdot \frac{\| g_u^m \|}{\sum_{u' \in \mathcal{U}_p} \| g_{u'}^m \|},
	\label{eq:directional_contribution}
\end{equation}
where $\cos(g_u^m, G^m)$ measures whether user $u$'s gradient direction is aligned with the aggregated update direction,
and $\| g_u^m \|$ reflects the contribution of user $u$ to modality $m$.

Based on this metric, we define the user set $\mathcal{U}_v$ and the user set $\mathcal{U}_t$ as the top-$K$ users ranked by $c_u^v$ and $c_u^t$, respectively. To measure the overlap between these two sets, we compute their Jaccard similarity:
\begin{equation}
J(\mathcal{U}_v, \mathcal{U}_t)
=
\frac{|\mathcal{U}_v \cap \mathcal{U}_t|}{|\mathcal{U}_v \cup \mathcal{U}_t|}.
\end{equation}

\begin{figure}[!t]
	\centering
	\includegraphics[scale=0.5]{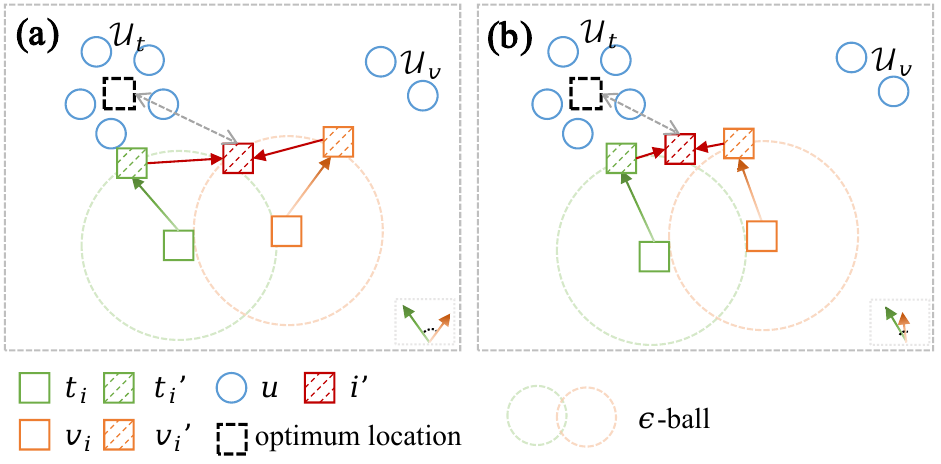} 
	\caption{Illustration of objective inconsistency across modalities. (a) Visual and textual perturbations are dominated by user groups $\mathcal{U}_v$ and $\mathcal{U}_t$, leading to conflicting promotion directions after multimodal fusion. (b) With alignment loss, the perturbations from different modalities are constrained to align in a consistent direction, enabling the fused embedding to effectively reach the optimum location.} %
	\label{Fig-mismatch}
\end{figure}

A low Jaccard similarity indicates that the users dominating the visual and textual perturbation updates are largely different, suggesting that the two modalities are implicitly driven by different user groups and, consequently, toward different effective optimization objectives.
As illustrated in Fig.~\ref{Fig-mismatch}(a), the visual perturbation $\Delta_v^i$ is primarily driven by users in $\mathcal{U}_v$, while the textual perturbation $\Delta_t^i$ is dominated by users in $\mathcal{U}_t$. When such modality-specific perturbations are subsequently fused by the MRS, the resulting item representation fails to simultaneously benefit all users, leading to suboptimal promotion outcomes.

\begin{figure}[tbp]
	\centering
	\subfigure[VBPR-Baby]{\includegraphics[width=0.48\columnwidth]{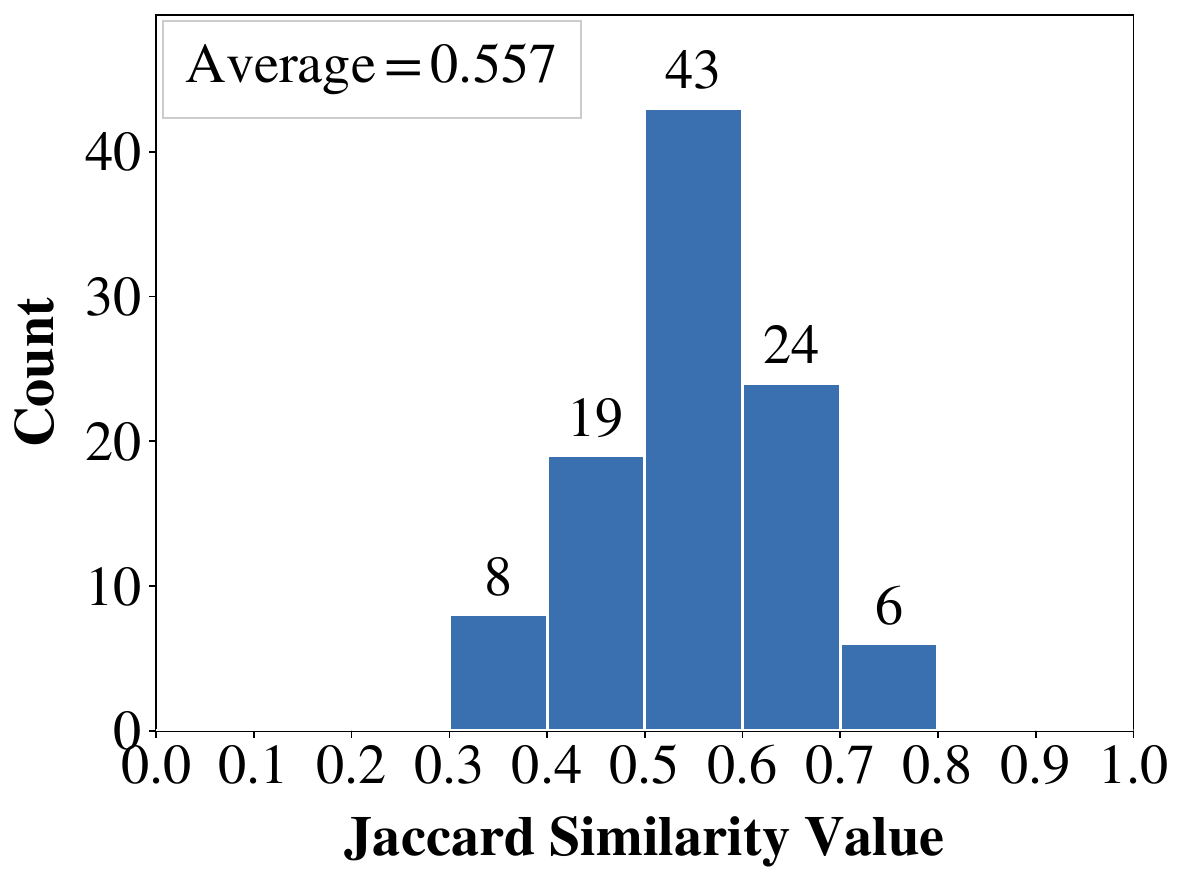}}\hfil
	\subfigure[VBPR-Sports]{\includegraphics[width=0.48\columnwidth]{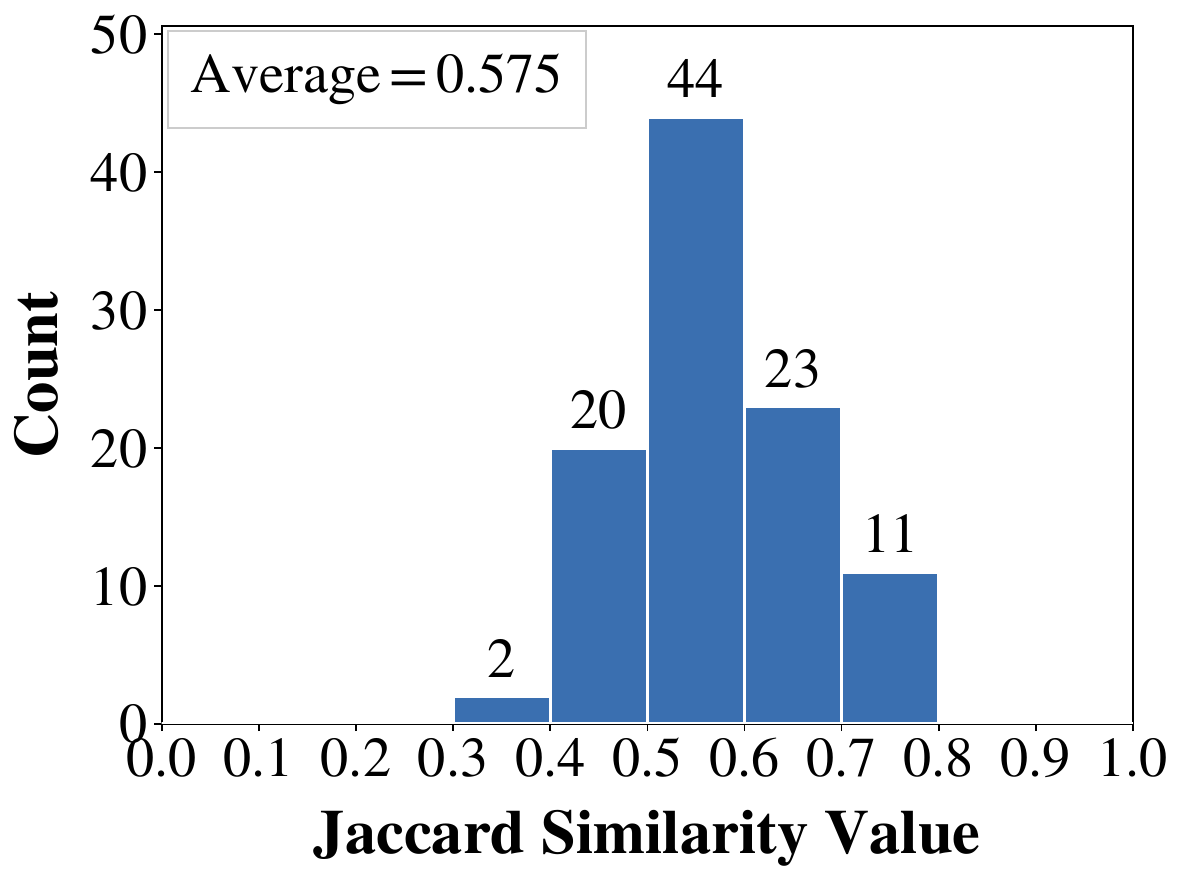}}
	\caption{Distribution of Jaccard Similarity between $\mathcal{U}_v$ and $\mathcal{U}_t$.}
	\label{Fig-jaccard_distribution}
\end{figure}

\subsection{Empirical Evidence of Cross-modal Gradient Mismatch}


As mentioned in Section \ref{sec:introduction}, we conduct vanilla FGSM-based promotion attacks on the VBPR model across three datasets. For each dataset, we randomly sample 100 unpopular items whose interaction counts are no greater than 5. For each targeted item, we identify the user sets $\mathcal{U}_v$ and $\mathcal{U}_t$ using the directional contribution metric in Equation~(\ref{eq:directional_contribution}), and conduct the promotion attack variants by constructing the promotion loss over  $\mathcal{U}_p$, $\mathcal{U}_v$ and $\mathcal{U}_t$, respectively.

The results indicate that aligning the attack objective with modality-specific user groups is crucial, as visual and textual perturbations are driven by distinct user subsets. Specifically, Figure \ref{Fig-jaccard_distribution} reveals a consistently low Jaccard similarity between  $\mathcal{U}_v$ and  $\mathcal{U}_t$, confirming their limited overlap. This distinctness is further corroborated by Figure \ref{Fig-motivation}, which demonstrates that optimizing over $\mathcal{U}_v$ or $\mathcal{U}_t$ yields significantly higher attack gains than using $\mathcal{U}_p$, as aggregating heterogeneous groups tends to dilute modality-specific optimization signals. (See Appendix \ref{Appendix:case_study} for a detailed case study visualizing this phenomenon on a specific item.)

In summary, cross-modal gradient mismatch limits the effectiveness of multimodal perturbations. Since effective defense requires training against worst-case attacks, it is critical to harmonize optimization directions to maximize attack potency. This motivates our UAT-MC framework, which uses gradient alignment to generate stronger adversarial examples, thereby driving the model to learn more robust representations.

\begin{figure*}[t]
	\centering
	\includegraphics[scale=0.65]{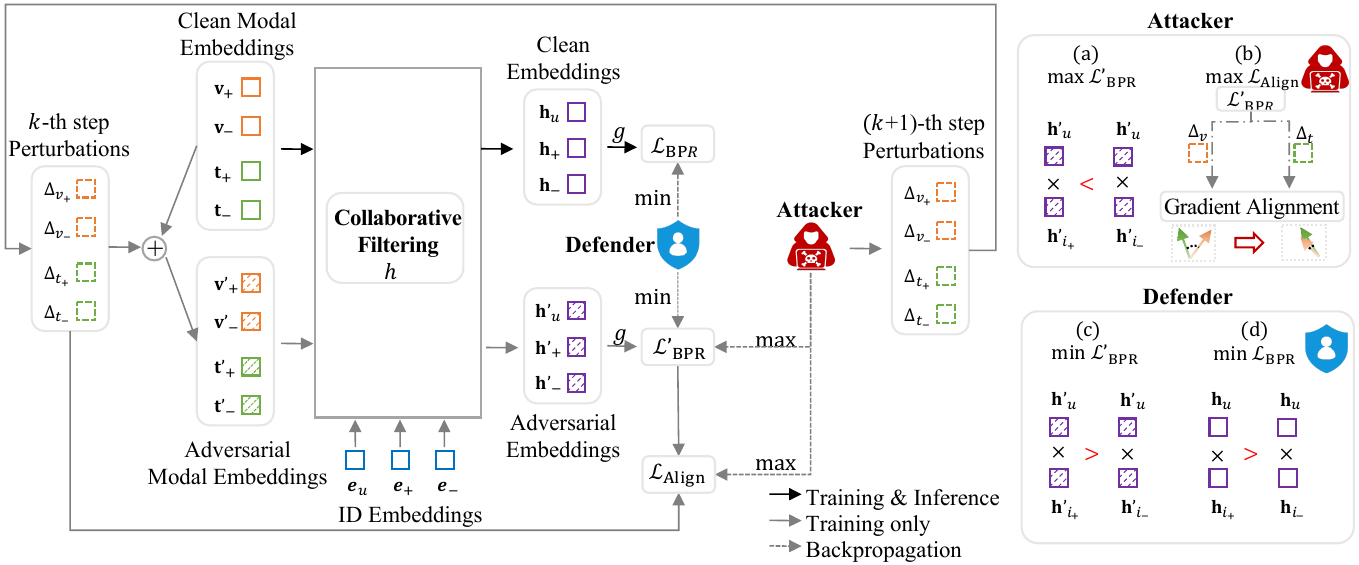}
	\caption{The framework of UAT-MC.}
	\label{fig:PromotionAttack}
\end{figure*}
\section{Methodology}
    \label{defend}
Figure \ref{fig:PromotionAttack} shows the overview of UAT-MC. In Figure \ref{fig:PromotionAttack}, the targeted MRS is divided into two parts, i.e., $f = g \circ h$, where $h$ is the encoder for collaborative filtering and $g$ is the decoder. Usually, $h$ is implemented as a GNN \cite{wei_2019_MMGCN} or an MLP \cite{he_2016_VBPR} for fusion of multimodal embeddings, while $g$ is implemented as an inner product. A training instance consists of user ID embedding $\mathbf{e}_u$, a positive item's ID embedding $\mathbf{e}_+$, a negative item's ID embedding $\mathbf{e}_-$, the positive item's modality embeddings $\mathbf{v}_+$, $\mathbf{t}_+$, and the negative item's modality embeddings $\mathbf{v}_-$ and $\mathbf{t}_-$. 

During the $k$-th iteration, UAT-MC first generates the adversarial modality embeddings $\mathbf{v}'_+$, $\mathbf{t}'_+$, $\mathbf{v}'_-$, and $\mathbf{t}'_-$, by adding the perturbations generated in the last step. Then through the collaborative filtering encoder $h$, UAT-MC produces the user embedding $\mathbf{h}_u$ by fusing with the multimodal content of the items interacted with $u$, the positive and negative item embeddings $\mathbf{h}_+$ and $\mathbf{h}_-$ by fusing their respective multimodal embeddings, and their respective adversarial sample embeddings $\mathbf{h}'_u$, $\mathbf{h}'_+$ and $\mathbf{h}'_-$. Based on these embeddings, the decoder $g$ computes the ranking losses $\mathcal{L}_{\text{BPR}}$ and $\mathcal{L}'_{\text{BPR}}$ on the clean embeddings and the adversarial embeddings, respectively. On these losses, UAT-MC conducts a min-max game to improve the robustness of the MRS $f$.

It is worth noting that to achieve worst-case adversarial robustness for $f$, UAT-MC promotes consistent perturbation directions across different modalities by maximizing the gradient alignment loss $\mathcal{L}_{\text{Align}}$ when generating adversarial perturbations. This approach realizes the multimodal coordination, thereby enhancing the aggressiveness of adversarial examples.

\subsection{Untargeted Adversarial Training}

As mentioned in Section \ref{sec:introduction}, in evasion attacks, the specific targeted item at inference time is unknown during the training phase. To address this challenge, we propose untargeted an adversarial training for MRSs, which treats all items as potential targets of evasion-based promotion attacks. This approach is motivated by the fact that recommendation can be formulated as a multi-label classification task, where each item serves as a distinct label. Untargeted adversarial training improves the robustness of the model by strengthening the decision boundary against perturbations from arbitrary directions, thereby effectively defending against perturbations from any direction.

Specifically, given a user $u$, the attacker adds perturbations $[\Delta_{v_{+}},\Delta_{t_{+}}]$ to the positive item's multimodal embeddings $[\mathbf{v}_{+},\mathbf{t}_{+}]$ and the perturbations $[\Delta_{v_{-}},\Delta_{t_{-}}]$ to the negative item's multimodal embeddings$[\mathbf{v}_{-},\mathbf{t}_{-}]$, which result in the adversarial modality embeddings $\mathbf{v}'_+$, $\mathbf{t}'_+$, $\mathbf{v}'_-$, and $\mathbf{t}'_-$. Then the final adversarial embeddings are computed as:
\begin{equation}
\mathbf{\textbf{h}}'_u, \mathbf{\textbf{h}}'_{+}, \mathbf{\textbf{h}}'_{-} = h(R,\mathbf{e}_{u}, \mathbf{e}_{+},\mathbf{e}_{-}, \mathbf{v}'_{+}, \mathbf{v}'_{-}, \mathbf{t}'_{+}, \mathbf{t}'_{-}).
\label{eq:MRS_adversarial_embedding}
\end{equation}

Then the untargeted adversarial training is defined as:
\begin{equation}
\min_{\Theta} \max_{\Delta_v,\Delta_t} \mathcal{L}'_\text{BPR}(\Theta, \Delta_v,\Delta_t),
\label{eq:UAT}
\end{equation}
where $\mathcal{L}'_\text{BPR}(\Theta, \Delta_v,\Delta_t) $ is defined as:

\begin{equation}
\mathcal{L}'_\text{BPR}(\Theta, \Delta_v,\Delta_t) = \sum_{(u,i_{+},i_{-}) \in \mathcal{D}} -\ln \sigma(\textbf{h}'_{u} \cdot \textbf{h}'_{+}  -\textbf{h}'_{u} \cdot \textbf{h}'_{-}).
\label{eq:adversarial_BPR}
\end{equation}

\subsection{Multimodal Coordination}

As analyzed in Section 3, independently perturbing each modality in multi-user promotion settings often results in cross-modal gradient mismatch, reducing the effectiveness of adversarial attacks.
While identifying and optimizing over specific user subsets (i.e., $\mathcal{U}_v$ and $\mathcal{U}_t$) can improve attack performance, it requires computing user-wise gradients for the entire user base, incurring a prohibitively high computational cost.
To circumvent this bottleneck, we propose a lightweight \textbf{multimodal coordination} mechanism. 
Instead of explicitly partitioning users, we focus on directly rectifying the divergent optimization directions in the gradient space. 
By enforcing gradient-level alignment between modalities, we effectively mitigate the mismatch without the need for costly user identification. 
This approach achieves coordination with negligible computational overhead, requiring only minimal additional operations during backpropagation.

We first compute the gradients of the adversarial loss with respect to the perturbations in each modality:
\begin{equation}
	\Gamma_v = \nabla_{\Delta_v} \mathcal{L}'_{\text{BPR}},\Gamma_t = \nabla_{\Delta_t} \mathcal{L}'_{\text{BPR}}.
\end{equation}

Then, we define the alignment loss as the sum of the cosine similarities between the gradients of visual and textual modalities for both positive and negative items:
\begin{equation}
	\mathcal{L}_{\text{Align}} = cos(\Gamma_{v_{+}},\Gamma_{t_{+}}) + cos(\Gamma_{v_{-}},\Gamma_{t_{-}}).
	\label{eq: align_loss}
\end{equation}

Maximizing $\mathcal{L}_{\text{Align}}$ encourages the perturbations in different modalities to be directionally aligned, thereby pushing the targeted item toward the adversarial region.

\subsection{Optimization}

The training is divided into two phases:
\begin{itemize}
	\item \textbf{Attacking} (Max phase): Generate the most disruptive perturbations $\Delta_v$ and $\Delta_t$ within $\ell_2$-norm budgets $\epsilon_v$ and $\epsilon_t$ by maximizing $ \lambda \mathcal{L}'_{\text{BPR}} + \alpha \mathcal{L}_{\text{Align}}$.
	
	\item \textbf{Defending} (Min phase): Update model parameters $\Theta$ by minimizing $\mathcal{L}_{\text{min}} = \mathcal{L}_{\text{BPR}} + \lambda \mathcal{L}'_{\text{BPR}}(\Delta_v, \Delta_t) + \beta \|\Theta\|_2$.
\end{itemize}
The overall training process is described in Algorithm~\ref{alg:UAT-MC}, where the MRS is pre-trained using the standard BPR loss.

\begin{algorithm}[t]
    \caption{UAT-MC}
    \label{alg:UAT-MC}
    \begin{algorithmic}[1] 
        \Require Training data $\mathcal{D}$; Learning rate $\eta$; Perturbation budgets $\epsilon_t$, $\epsilon_v$; Hyperparameters $\lambda$, $\alpha$, $\beta$
        \Ensure Robust MRS model parameters $\Theta$
        \State Initialize $\Theta$ from normal-trained MRS
        \While{not converged}
            \State Randomly draw an example $(u,i_{+},i_{-})$ from $\mathcal{D}$
            
            \State \textit{// Max phase}
            \State $\mathcal{L}'_\text{BPR} = -\ln \sigma(\textbf{h}'_{u} \cdot \textbf{h}'_{+} - \textbf{h}'_{u} \cdot \textbf{h}'_{-})$
            \State $\Gamma_v = \nabla_{\Delta_v} \mathcal{L}'_{\text{BPR}}$, $\Gamma_t = \nabla_{\Delta_t} \mathcal{L}'_{\text{BPR}}$
            \State $\mathcal{L}_{\text{Align}} = \cos(\Gamma_{v_{+}},\Gamma_{t_{+}}) + \cos(\Gamma_{v_{-}},\Gamma_{t_{-}})$
            \State $\mathcal{L}_{\text{max}} = \mathcal{L}'_{\text{BPR}} + \alpha \mathcal{L}_{\text{Align}}$
            
            \State $\Delta_v \leftarrow \epsilon_v \cdot \frac{\nabla_{\Delta_v} \mathcal{L}'_{\text{max}}}{\| \nabla_{\Delta_v} \mathcal{L}'_{\text{max}} \|_2}$, $\Delta_t \leftarrow \epsilon_t \cdot \frac{\nabla_{\Delta_t} \mathcal{L}'_{\text{max}}}{\| \nabla_{\Delta_t} \mathcal{L}'_{\text{max}} \|_2}$
            
            \State \textit{// Min phase}
            \State $\mathcal{L}_{\text{min}} = \mathcal{L}_{\text{BPR}} + \lambda \mathcal{L}'_{\text{BPR}}(\Delta_v, \Delta_t) + \beta \|\Theta\|_2$
            \State $\Theta \leftarrow \Theta - \eta \nabla_{\Theta} \mathcal{L}_{\text{min}}$
        \EndWhile
        \State \Return $\Theta$
    \end{algorithmic}
\end{algorithm}

    \begin{table}
    	\centering
    	\small
    	\begin{tabular}{|c|c|c|c|c|}
    		\hline
    		\textbf{Dataset} & \textbf{\#Users} & \textbf{\#Items} & \textbf{\#Interactions} & \textbf{Sparsity} \\ \hline
    		Baby             & 19,445           & 7,037            & 160,792                 & 99.883\%          \\ \hline
    		Sports           & 35,598           & 18,357           & 296,337                 & 99.955\%          \\ \hline
    		Clothing        & 39,387           & 23,033           & 278,677                 & 99.969\%          \\ \hline
    	\end{tabular}
    	\caption{Statistics of the three experimental datasets.}
    	\label{table:dataset}
    \end{table}

\section{Experiments}
    \label{sec:experiment}

\input{Tables/table_defense_result}

The objectives of experiments are to answer the following research questions:
\begin{itemize}
	\item \textbf{RQ1:} Can UAT-MC defend against evasion-based promotion attacks?
	\item \textbf{RQ2:} Does multimodal coordination enhance the effectiveness of adversarial defense?
	\item \textbf{RQ3:} How does the perturbation budget in adversarial training affect the defense performance?
	\item \textbf{RQ4:} How do the hyper-parameters affect the trade-off between recommendation performance and adversarial robustness?
\end{itemize}

All experiments are implemented using PyTorch 2.2.0 and run on an NVIDIA RTX 4090 GPU with 48GB of memory.

\subsection{Experimental Settings}
\subsubsection{Datasets} 
We conduct experiments on three Amazon datasets Baby, Sports and Clothing \cite{he_2016_ups_Amazon}, which have recently been adopted in MRSs research \cite{yu_2023_multi_MGCN,guo_2024_lgmrec_LGMRec,li_2025_Teach_GUIDER}. The dataset statistics are summarized in Table \ref{table:dataset}. In our work, we use the pre-extracted visual features and textual features that have been published in \cite{zhou_2023_bootstrap_BM3,Zhou_2023_A_MMRec}.

\begin{figure*}[htbp]
	\centering
	\includegraphics[scale=0.5]{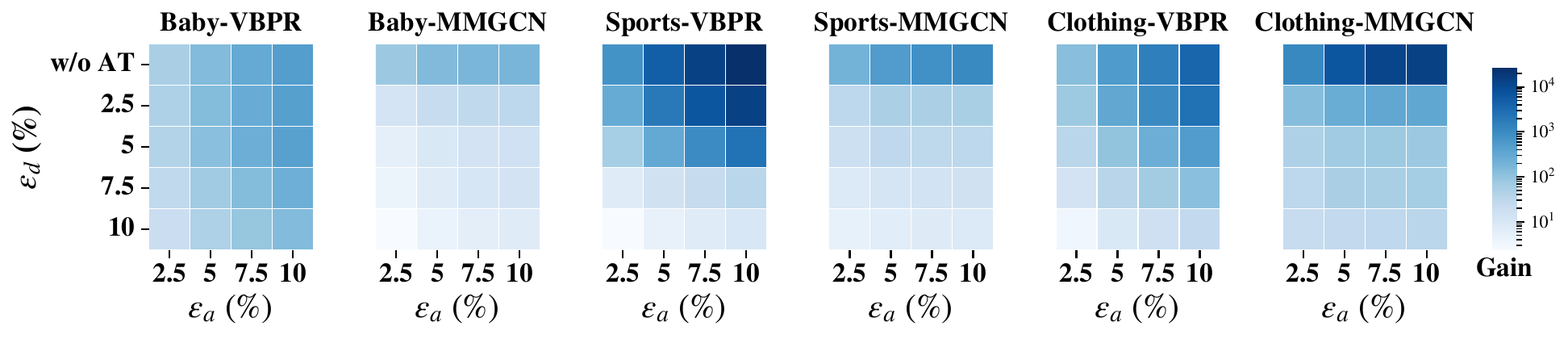}
	\caption{Visualization of attack effectiveness under varying budgets. The heatmaps display the $Gain_{Hit@50}$ (log scale) across three datasets and two models. The x-axis represents the attack budget $\epsilon_a$, and the y-axis represents the defense budget $\epsilon_d$. Darker colors indicate higher attack gains (weaker defense), while lighter colors indicate effective defense.}
	\label{fig:diff_epsilon_fgsm_align}
\end{figure*}

\subsubsection{Evaluation Metrics}
To measure recommendation performance before and after adversarial training, we use Recall$@10$ and NDCG$@10$ under the leave-one-out evaluation protocol following standard evaluation settings in prior works \cite{zhou_2023_bootstrap_BM3,guo_2024_lgmrec_LGMRec}. To evaluate the effectiveness of the promotion attack, we adopt the hit rate $\text{Hit}_{i}@K={N^i_{rec}}/{|\mathcal{U}|}\cdot100\%$ and the relative improvement ratio $\text{Gain}_{\text{Hit}_{i}@K}$ $=(\text{Hit}_{\text{after}}@K $ $- \text{Hit}_{\text{before}}@K)/{\text{Hit}_{\text{before}}@K}\cdot100\%$ as evaluation metrics, where  $N^i_{\text{rec}}$ is the number of users for whom the targeted item $i$ appears in the top-$K$ recommendation list (with K defaulted to 50), $\text{Hit}_{\text{after}}@K$ and $\text{Hit}_{\text{before}}@K$ denote the hit rate of the targeted item after and before the attack, respectively. A larger $\text{Gain}_{\text{Hit}@50}$ indicates a more effective promotion attack, suggesting weaker model robustness. 

\subsubsection{Targeted Items} 
At the MRS's inference phase,  we randomly select 100 targeted items from unpopular items, whose interaction count is 5. We conduct promotion attacks on each targeted item individually and record the average $\text{Hit}_{i}@K$ and $\text{Gain}_{\text{Hit}_{i}@K}$.

\subsubsection{Victim Models}
We choose two mainstream and classical multimodal recommendation models as our victim models: 
\begin{itemize}
	\item VBPR \cite{he_2016_VBPR}: As a representative of MLP-based models, VBPR incorporates visual features for user preference learning with BPR loss. Following \cite{zhou_2023_bootstrap_BM3,guo_2024_lgmrec_LGMRec}, we concatenate the multimodal embeddings and the item ID embedding as the item embeddings.
	\item MMGCN \cite{wei_2019_MMGCN}: As a representative of GCN-based models, MMGCN constructs three modal-specific graph to learn different modality features. It concatenates all modality embeddings to obtain the representations of users or items.
\end{itemize}
We follow MMRec \cite{Zhou_2023_A_MMRec} to save the model parameters at the point of best performance. The hyper-parameters search spaces are provided in the Appendix~\ref{Appendix:hyperparameter_search_space}. 

\subsubsection{Promotion Attacks}  We conduct promotion attacks by maximizing $\mathcal{L}_{\text{promotion}}$ (Eq. \ref{eq:L_promotion}) using two standard strategies: FGSM \cite{goodfellow_2014_explaining_FGSM} and PGD \cite{madry2017towards_PGD}. FGSM serves as a single-step baseline, whereas PGD functions as a stronger iterative attacker. In our experiments, the PGD attack is configured with 10 iterations and a step size of $\alpha=1.25 \cdot \epsilon_m / 10$.
\begin{figure}[tbp]
	\centering
	\subfigure[MMGCN-Baby ($\lambda$)]{\includegraphics[width=0.45\columnwidth]{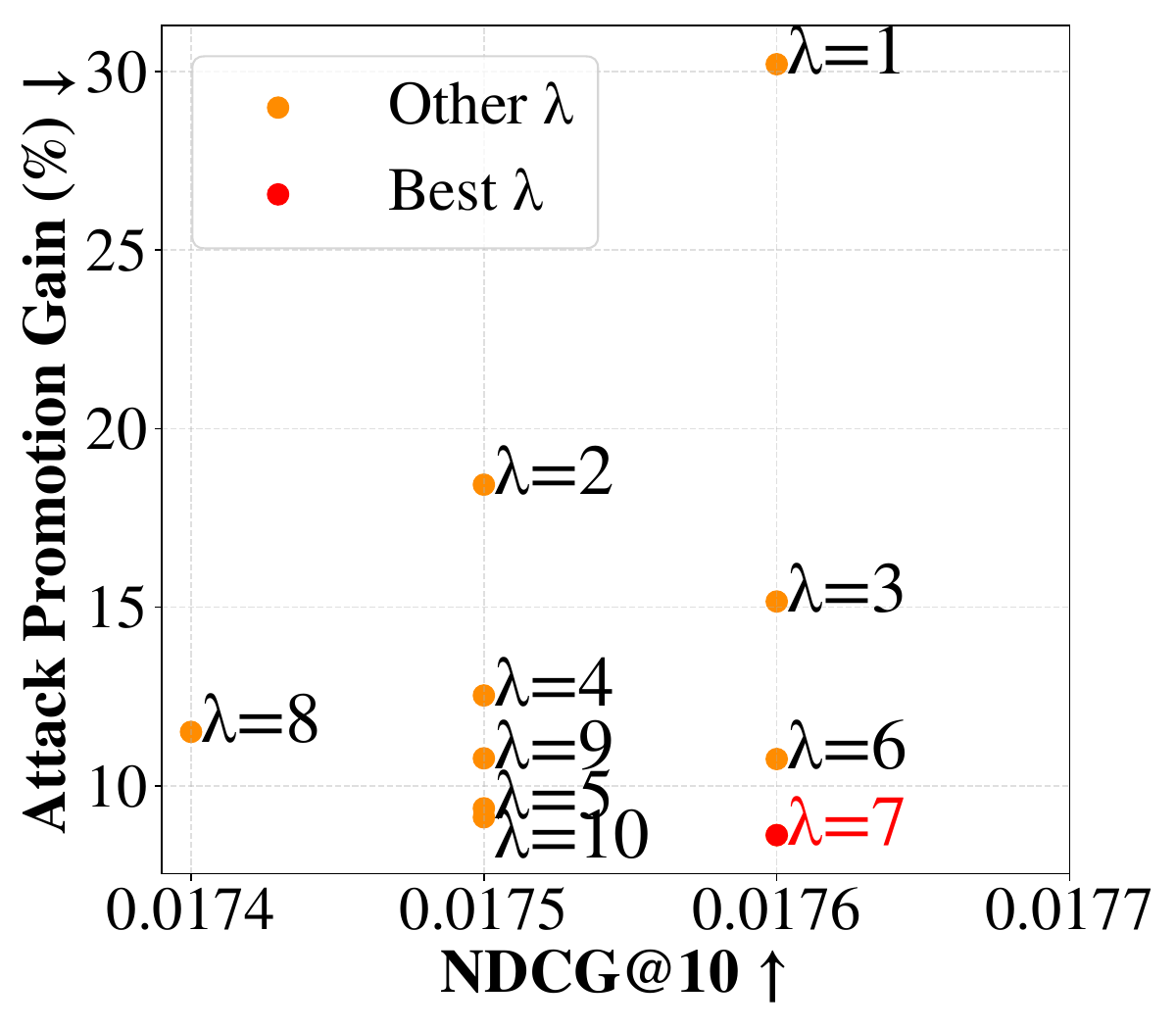}}\hfil
    \subfigure[MMGCN-Baby ($\alpha$)]{\includegraphics[width=0.45\columnwidth]{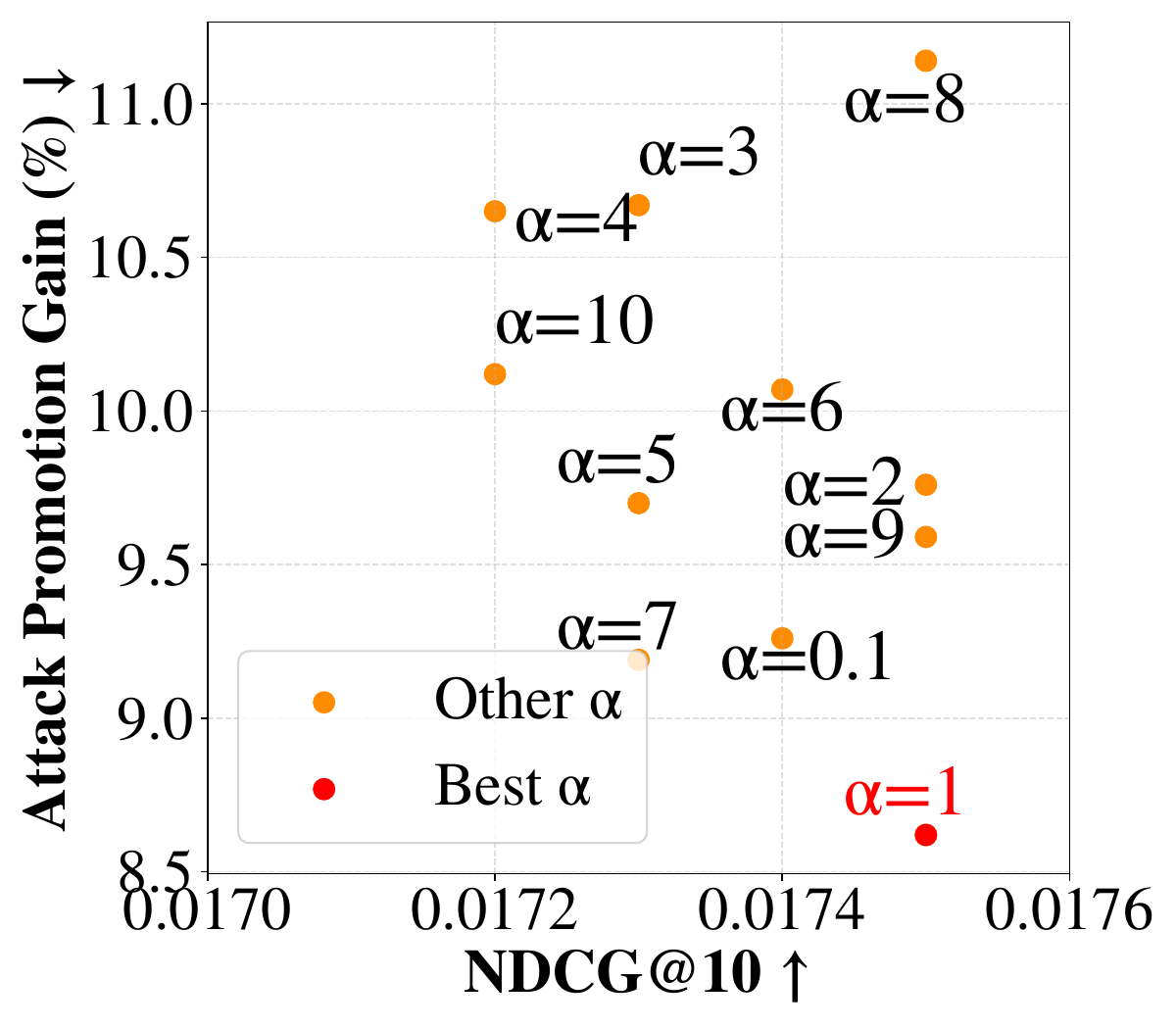}}
	\caption{Trade-off between recommendation performance (NDCG$@10$) and adversarial robustness ($\text{Gain}_{\text{Hit}@50}$) under PGD(w/ $\mathcal{L}_{\text{Align}}$)-based attack with varying $\lambda$ and $\alpha$ on the Baby dataset.}
	\label{fig:pareto_attack_lambda_and_alpha_baby}
\end{figure}

\subsubsection{Defenders} 

To verify the effectiveness of the Untargeted Adversarial Training and the Multimodal Coordination, we compare UAT-MC with two baseline models: one is the victim model without any defense mechanisms, denoted as \textbf{w/o AT}, and the other is the variant of UAT-MC, denoted as \textbf{UAT}, which applies perturbations to both visual and textual modalities independently without coordination.

\subsubsection{Implementation Details}
During adversarial training phase, we set the maximum perturbation magnitude $\epsilon_{m}$ as 10\% of the 2-norm of the input embedding for modality $m$, the coefficient $\alpha$ is searched in \{0.1,1,2,3,4,5,6,7,8,9,10\}, the coefficient $\lambda$ is searched in \{1,2,3,4,5,6,7,8,9,10\}.

\subsection{Defense Effect (RQ1 and RQ2)}

We conduct a systematic evaluation of the performance of different defense methods on two attack scenarios (FGSM-based and PGD-based) on two mainstream MRSs (VBPR \cite{he_2016_VBPR} and MMGCN \cite{wei_2019_MMGCN}). The victim models take the original and perturbed multimodal embeddings as input, respectively. Meanwhile, we also report the recommendation performance on the clean test set to reflect the impact of adversarial training on overall recommendation quality. The results are shown in Table \ref{table:defense_result}, from which we have the following observations:
\begin{itemize}
	\item \textbf{Across all dataset-model combinations, unpopular items consistently exhibit vulnerability to promotion attacks (w/o $\mathcal{L}_{\text{Align}}$)}, which risk misleading the target model into treating them as popular items. It can be observed that perturbations generated by PGD are more effective in promoting the targeted items compared to those generated by FGSM. Statistically, on the Baby dataset, the average $\text{Hit}@50$ of unpopular items increases from 0.667\% to 3.865\% under FGSM-based perturbations, and further to 4.111\% under PGD-based perturbations. This implies that more than 3.444\% of users will be misled into recommending an item (e.g., targeted item) that should not appear in the recommendation lists. 
	\item \textbf{Attacks equipped with $\mathcal{L}_{\text{Align}}$ consistently achieve higher promotion gains than their counterparts without alignment,} indicating that explicitly coordinating visual and textual perturbations leads to more effective promotion. This improvement is particularly evident for stronger attacks. For example, under PGD-based attacks on the Baby dataset, introducing $\mathcal{L}_{\text{Align}}$ increases the $\text{Gain}_{\text{Hit}@50}$ of VBPR from 516.15\% to 520.99\%, and that of MMGCN from 446.76\% to 450.30\%. Similar trends are observed on the Sports and Clothing datasets, where alignment consistently yields additional promotion gains across both VBPR and MMGCN. 

	\item \textbf{After untargeted adversarial training (UAT), the MRSs exhibit significantly enhanced resistance to evasion-based promotion attacks.} The results show that the effectiveness of both FGSM-based and PGD-based attacks drops significantly after UAT. In the following, we analyze based on the average $\text{Gain}_{\text{Hit}@50}$ results of the two attacks (w/ $\mathcal{L}_{\text{Align}}$). On the Baby dataset, UAT reduces the $\text{Gain}{\text{Hit}@50}$ of VBPR from 503.39\% to 288.83\%, achieving a 42.62\% relative reduction. For MMGCN, the value drops significantly from 315.72\% to 8.42\%, corresponding to a 97.33\% reduction. 
    Similar trends are observed on the Sports dataset, where VBPR and MMGCN experience reductions of 99.95\% and 98.92\%, respectively. On the Clothing dataset, UAT reduces  $\text{Gain}_{\text{Hit}@50}$ by 76.50\% for VBPR and 99.71\% for MMGCN. 

	\item \textbf{UAT with Multimodal Coordination can further enhance its defense capability.} According to our observations, the $\text{Gain}_{\text{Hit}@50}$ of MMGCN on the Sports dataset drops from 25.45\% to 19.14\%, while that of VBPR on the Clothing dataset decreases significantly from 898.38\% to 27.70\% after applying multimodal coordination. Notably, the improvement in robustness is achieved with minimal impact on recommendation performance.
\end{itemize}

\subsection{Hyper-parameter Study (RQ3 and RQ4)}
This section explores how the perturbation budget $\epsilon_m$ and the hyper-parameters $\lambda$ and $\alpha$ in adversarial training influence the trade-off between adversarial robustness and recommendation performance. 

\begin{itemize}
	  \item \textbf{Impact of $\epsilon_d$ and $\epsilon_a$} To examine the robustness and generalization ability of UAT-MC, we conduct adversarial training with different defense perturbation budgets $\epsilon_d \in \{2.5\%, 5\%, 7.5\%, 10\%\}$, and evaluate the trained models under FGSM-based promotion attacks with varying attack budgets $\epsilon_a \in \{2.5\%, 5\%, 7.5\%, 10\%\}$. 
The results are visualized in Figure~\ref{fig:diff_epsilon_fgsm_align}, where specific numerical values are detailed in the Appendix~\ref{Appendix:appendix_budgets}.
We observe a clear transition from dark blue to light blue/white as $\epsilon_d$ increases, indicating that our defense method effectively mitigates the promotion attack. 

    \item \textbf{Impact of $\lambda$ and $\alpha$} We investigate the trade-off between adversarial robustness and recommendation performance by varying $\lambda \in [1, 10]$ and $\alpha \in [0.1, 10]$. Figure~\ref{fig:pareto_attack_lambda_and_alpha_baby} (Baby dataset) and the Appendix~\ref{Appendix:hyperparameter} (others) visualize this relationship, plotting recommendation performance (NDCG@10) against attack gain ($\text{Gain}_{\text{Hit}@50}$). A clear trade-off is observed: higher recommendation performance typically comes at the cost of increased vulnerability. 

\end{itemize}

\section{Conclusion}
    \label{sec:conclusion}

    In this work, we address the vulnerability of MRSs to promotion attacks. Crucially, we verify the existence of cross-modal gradient mismatch in multi-user promotion settings and proposed UAT-MC to mitigate it via a novel gradient alignment regularization. Extensive experiments demonstrate the effectiveness of UAT-MC in defending against evasion-based promotion attacks.

\section*{Acknowledgments}
This work is supported by Natural Science Foundation of Sichuan Province under grant 2024NSFSC0516 and National Natural Science Foundation of China under grant 61972270.

\bibliographystyle{named}
\bibliography{ijcai26}

\clearpage
\appendix

\section{Related Work}
\label{sec:relatedwork}

\subsection{Multimodal Recommender Systems}
Multimodal recommender systems (MRSs) aim to enhance recommendation performance by integrating multiple modalities of items. For example, 
VBPR \cite{he_2016_VBPR} extends traditional Matrix Factorization by integrating multimodal representations with item ID embeddings. GNN-based methods such as MMGCN \cite{wei_2019_MMGCN}, GRCN \cite{Wei_2020_Graph_GRCN} and DualGNN \cite{Wang_2023_DualGNN} encode representations for each modality to capture user's modal-specific preferences. Recent studies have sought to improve recommendation performance using various auxiliary information, such as the knowledge graph \cite{sun_2020_multi_MMKGs_MMGAT}, user co-occurrence graph \cite{zhou2023enhancing_DRAGON}, item-item relation graph \cite{zhang_2021_mining_LATTICE}, self-supervised learning \cite{zhou_2023_bootstrap_BM3}, 
hypergraph \cite{guo_2024_lgmrec_LGMRec}, multimodal alignment \cite{Liu_2024_AlignRec}, spectral theory \cite{Ong_2025_Spectrum_SMORE}. However, while above research enhances the personalization by utilizing the auxiliary modal information, MRSs exhibit heightened vulnerability to the promotion attacks.

\subsection{Promotion Attacks and Defenses}

As targeted adversarial strategy, promotion attacks can be typically classified into two paradigms based on implementation mechanisms: \textbf{Poisoning-based promotion attacks} achieve their objectives by injecting fabricated user profiles into the training set. For example, GSPAttack \cite{Thanh_2023_GSPAttack} proposes a generative surrogate-based poisoning framework for graph-based recommender systems (RSs). Infmix \cite{Wu_2023_Influence_Infmix} designs an influence-based threat estimator and a distribution-agnostic user generator to craft imperceptible yet impactful fake users. CLeaR \cite{wang_2024_unveiling_CLeaR} proposes a dual-objective attack framework that exploits representation dispersion and rank promotion, revealing the risks introduced by contrastive learning. Our work focuses on \textbf{evasion-based promotion attacks} that achieve the objectives at the inference phase by injecting imperceptible adversarial perturbations into the item's content.  For visually-aware recommendation models, TAaMR \cite{Di_2020_Targeted_TAaMR}, applies FGSM and PGD attacks on pre-trained visual encoders to misclassify the targeted item into a popular item' category. AIP \cite{Liu_2021_Adversarial_AIP_Image}, IPDGI \cite{Chen_2024_Visually_IPDGI}, IPDGI \cite{Chen_2024_Visually_IPDGI} and SPAF \cite{Yang_2024_Attacking_SPAF} generate imperceptible and diverse adversarial images to increase the similarity between the targeted item and popular items. RecTextAttack \cite{zhang_2024_stealthy_RecTextAttack} and TextSimu \cite{Wang_2024_llmpowered_TextSimu} use LLMs to manipulate item text content to boost their ranking in text-aware recommendation models, revealing new vulnerabilities introduced by large language models (LLMs).

To date, many studies have focused on defending against such attacks through \textbf{data filtering} or \textbf{robust training}. \textbf{Data filtering} methods aim to detect and remove adversarially manipulated items or inputs before they impact the recommendation model. For example, Re-writing Defense \cite{zhang_2024_stealthy_RecTextAttack} utilizes GPT-3.5-turbo to rewrite the adversarial text to defense LLM-based RSs. RecMR \cite{hsiao_2022_unsupervised_RecMR} uses an AutoEncoder as a detection encoding model to distinguish anomalies. NFGCN-TIA \cite{wang_2022_detecting_NFGCN-TIA} employs a GCN model to detect malicious users. \textbf{Robust training} enhances the model’s resilience by incorporating adversarial examples during training or modifying the learning objectives to reduce sensitivity to perturbations. APT \cite{wu_2021_Fight_APT}  simulates the poisoning process by injecting fake user data to foster a more robust system. AutoDenoise \cite{lin2023autodenoise} addresses the challenge of highly dynamic data distributions by employing a deep RL-based framework.

However, to the best of our knowledge, existing studies on adversarial robustness have primarily focused on single modal  recommender systems, without considering the joint presence of visual and textual modalities. From an adversarial perspective, launching attacks on both modalities simultaneously to achieve effective promotion is natural. This highlights the urgent need to investigate promotion attacks and defense specifically tailored to MRSs.

\section{Experiments}
\label{Appendix:Experiments}
\subsection{Hyper-parameter Study} 
\subsubsection{MRSs Training}
\label{Appendix:hyperparameter_search_space}
\input{Tables/tables_rs_hyperparameter_search_space.tex}
We follow MMRec \footnote{https://github.com/enoche/MMRec} to save the model parameters at the point of best performance. The hyper-parameters search spaces are provided in the Table \ref{Table:hyper_parameter}.

\subsubsection{Impact of $\lambda$ and $\alpha$}
\label{Appendix:hyperparameter}
\begin{figure}[htbp]
    \centering
    \subfigure[VBPR-Baby ($\lambda$)]{\includegraphics[width=0.45\columnwidth]{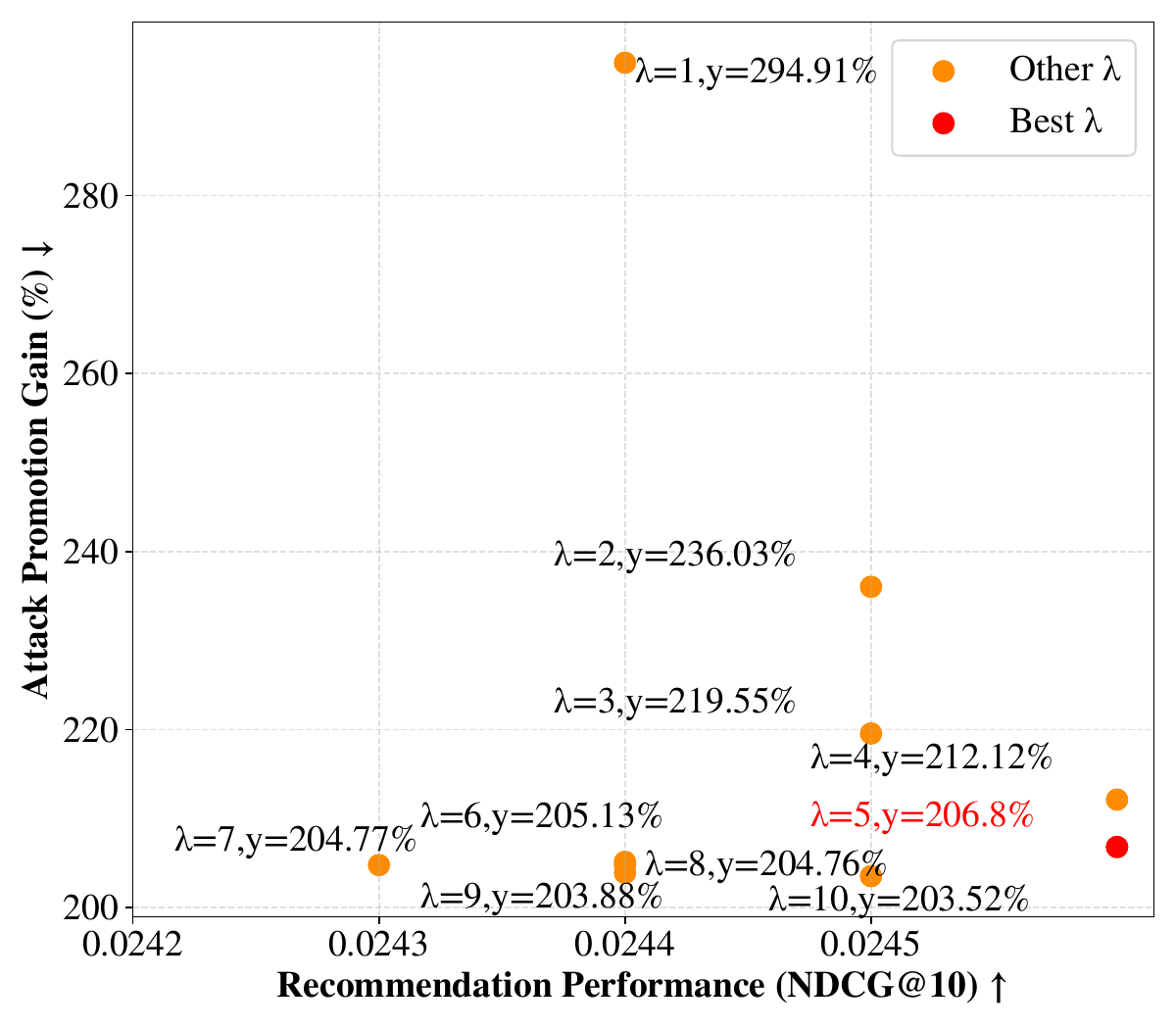}}\hfil
    \subfigure[VBPR-Baby ($\alpha$)]{\includegraphics[width=0.45\columnwidth]{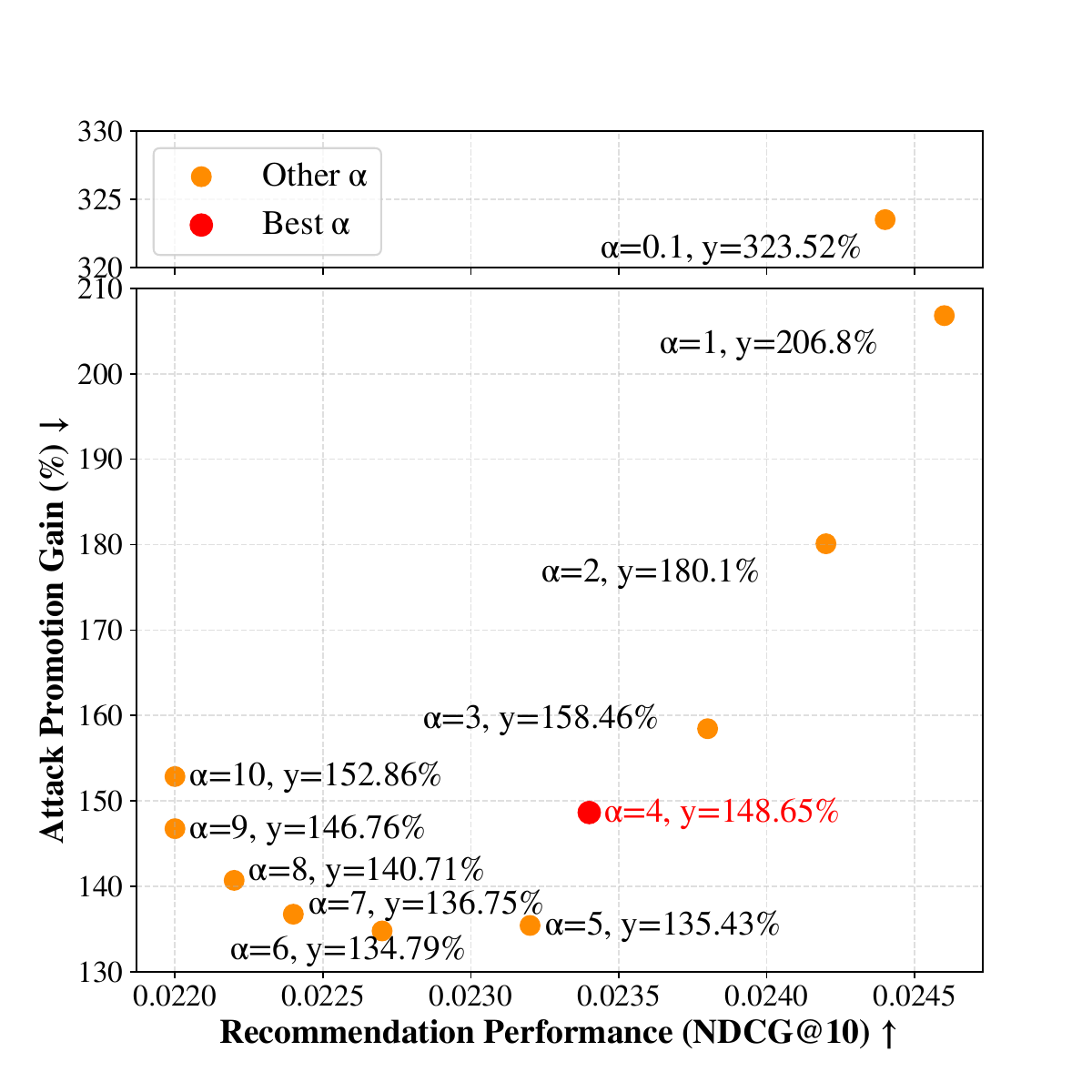}} \\
    
    \subfigure[VBPR-Sports ($\lambda$)]{\includegraphics[width=0.45\columnwidth]{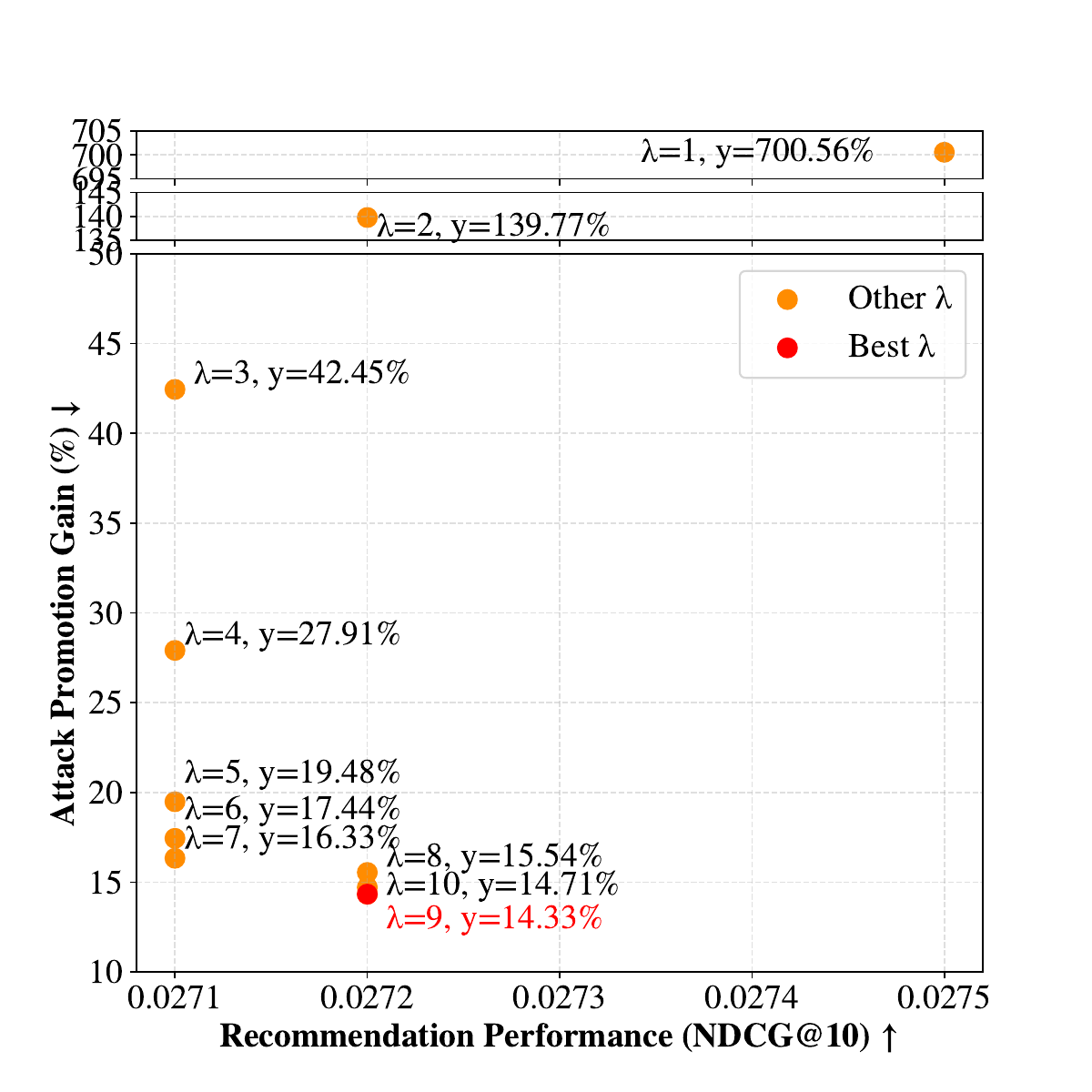}}\hfil
    \subfigure[VBPR-Sports ($\alpha$)]{\includegraphics[width=0.45\columnwidth]{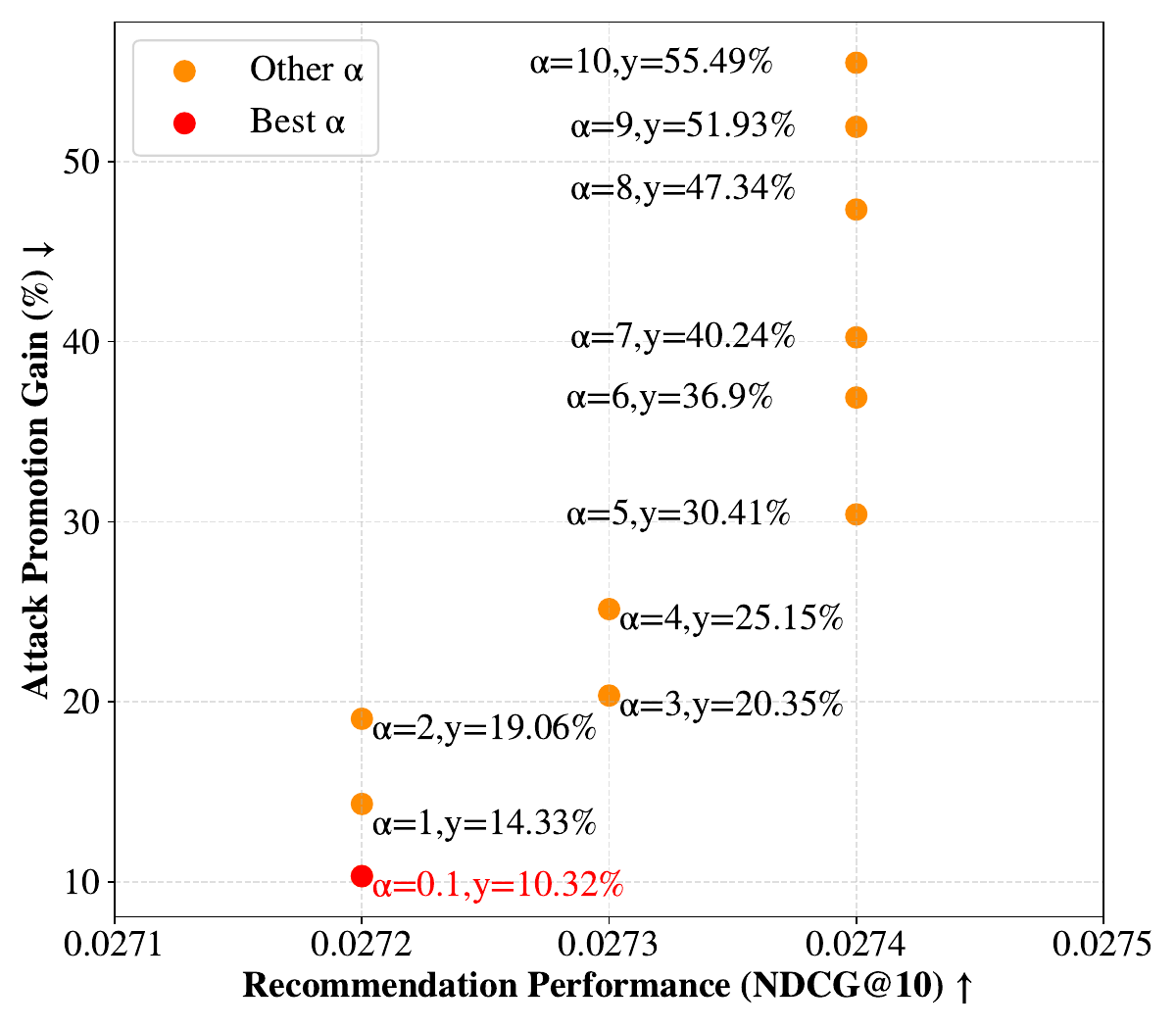}} \\
    
    \subfigure[VBPR-Clothing ($\lambda$)]{\includegraphics[width=0.45\columnwidth]{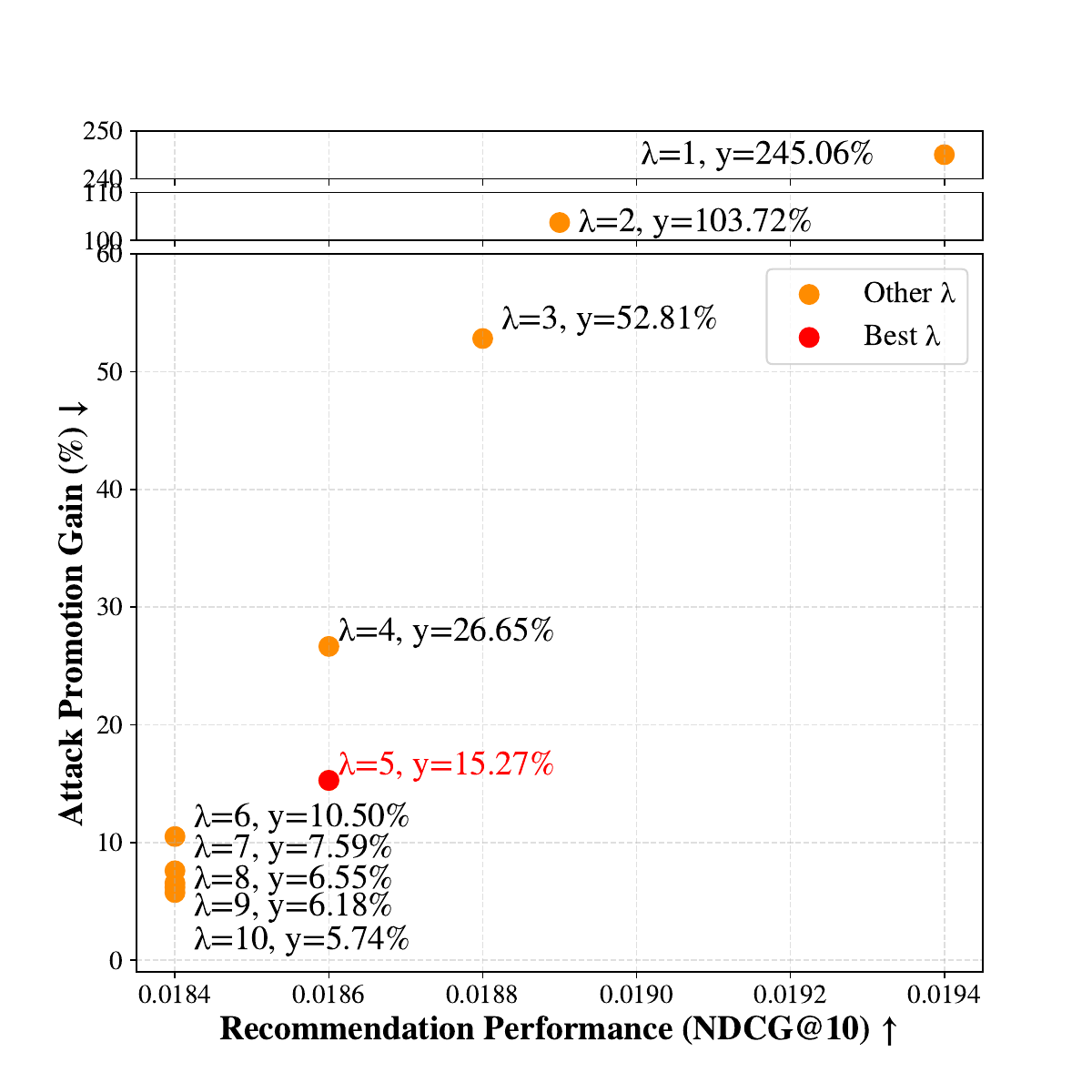}}\hfil
    \subfigure[VBPR-Clothing ($\alpha$)]{\includegraphics[width=0.45\columnwidth]{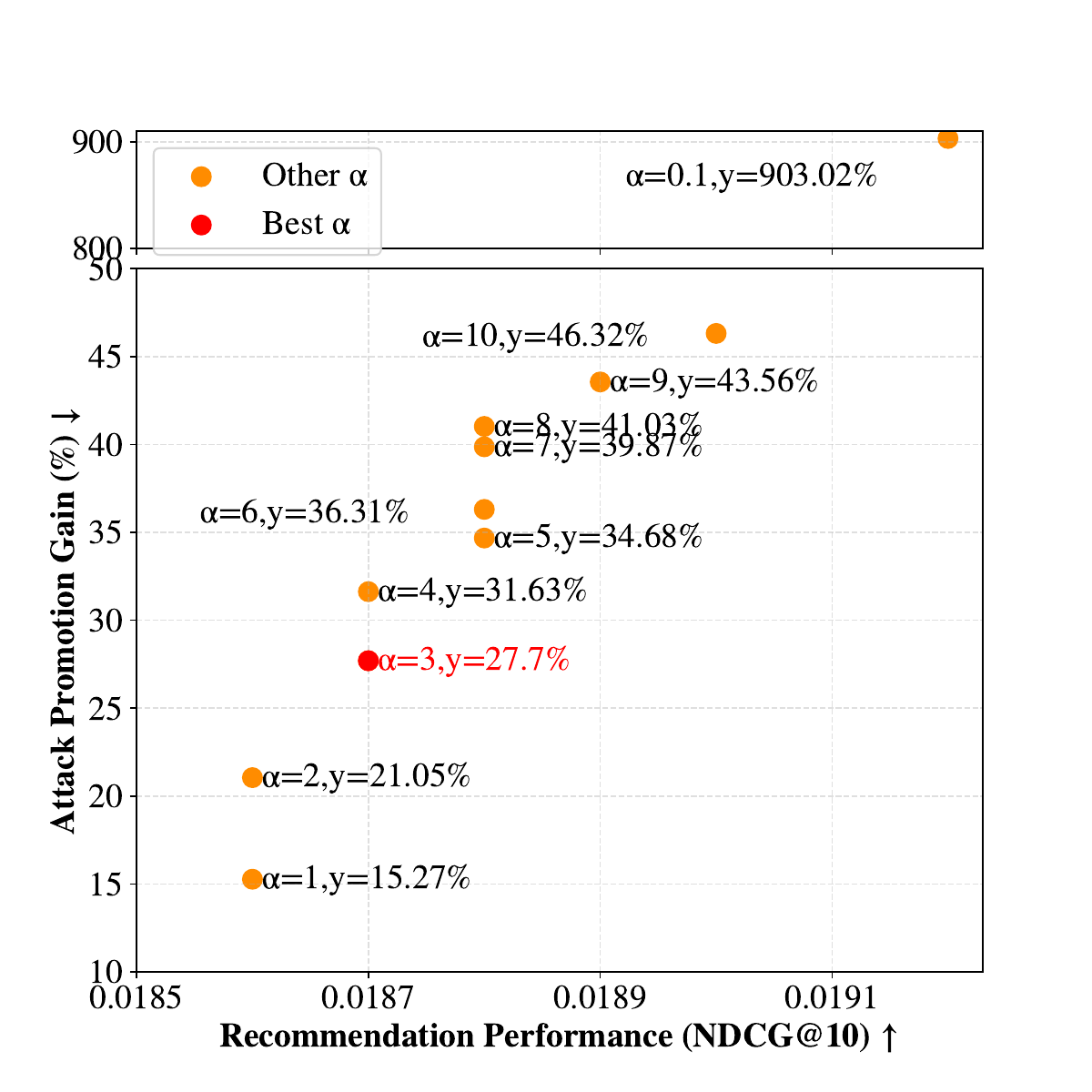}}

    \caption{Impact of $\lambda$   and $\alpha$  on the trade-off between accuracy and attack effectiveness for VBPR under PGD-based attack. }
    \label{fig:VBPR_PGD_Analysis}
\end{figure}

\begin{figure}[htbp]
    \centering
    \subfigure[MMGCN-Baby ($\lambda$)]{\includegraphics[width=0.45\columnwidth]{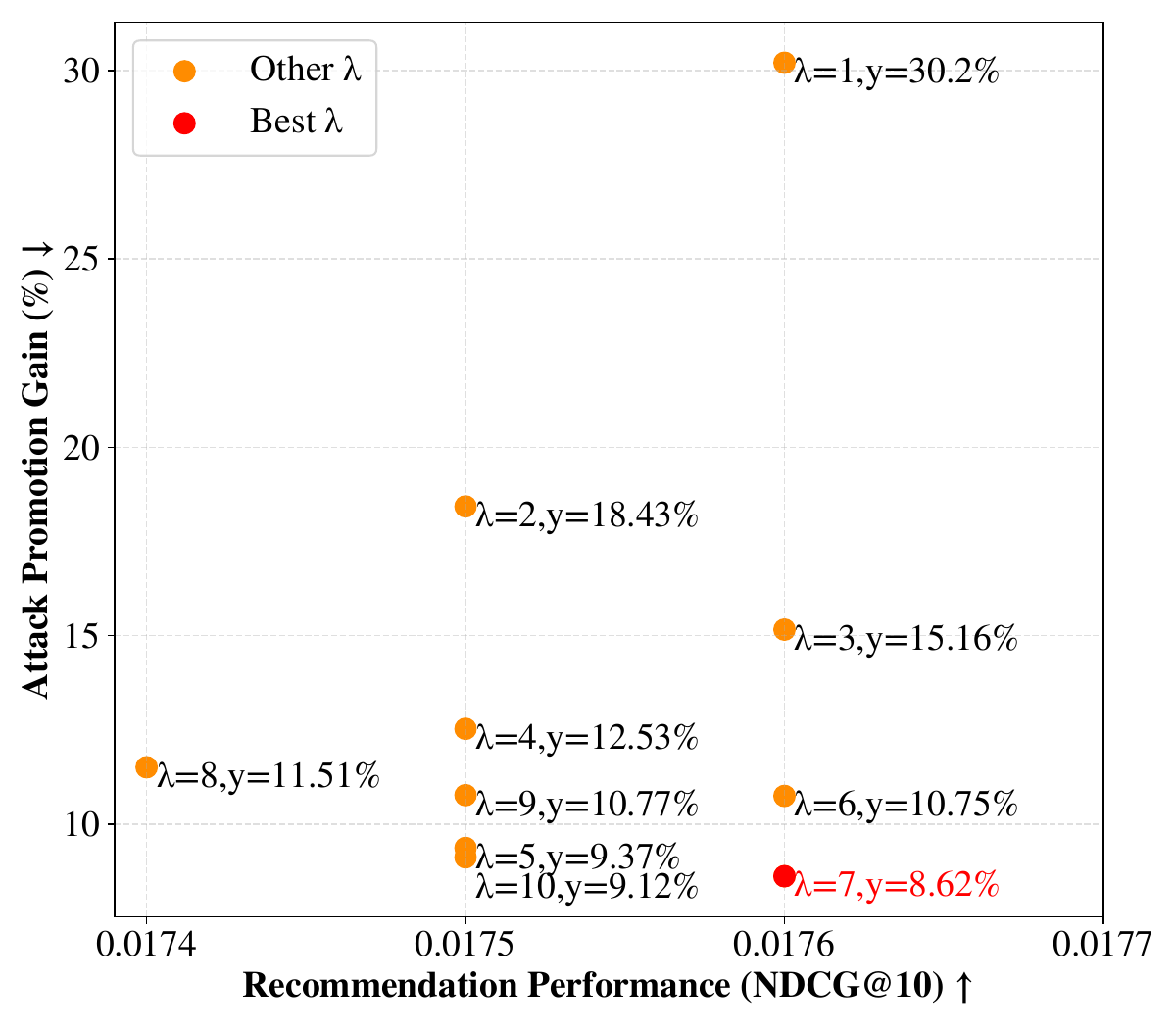}}\hfil
    \subfigure[MMGCN-Baby ($\alpha$)]{\includegraphics[width=0.45\columnwidth]{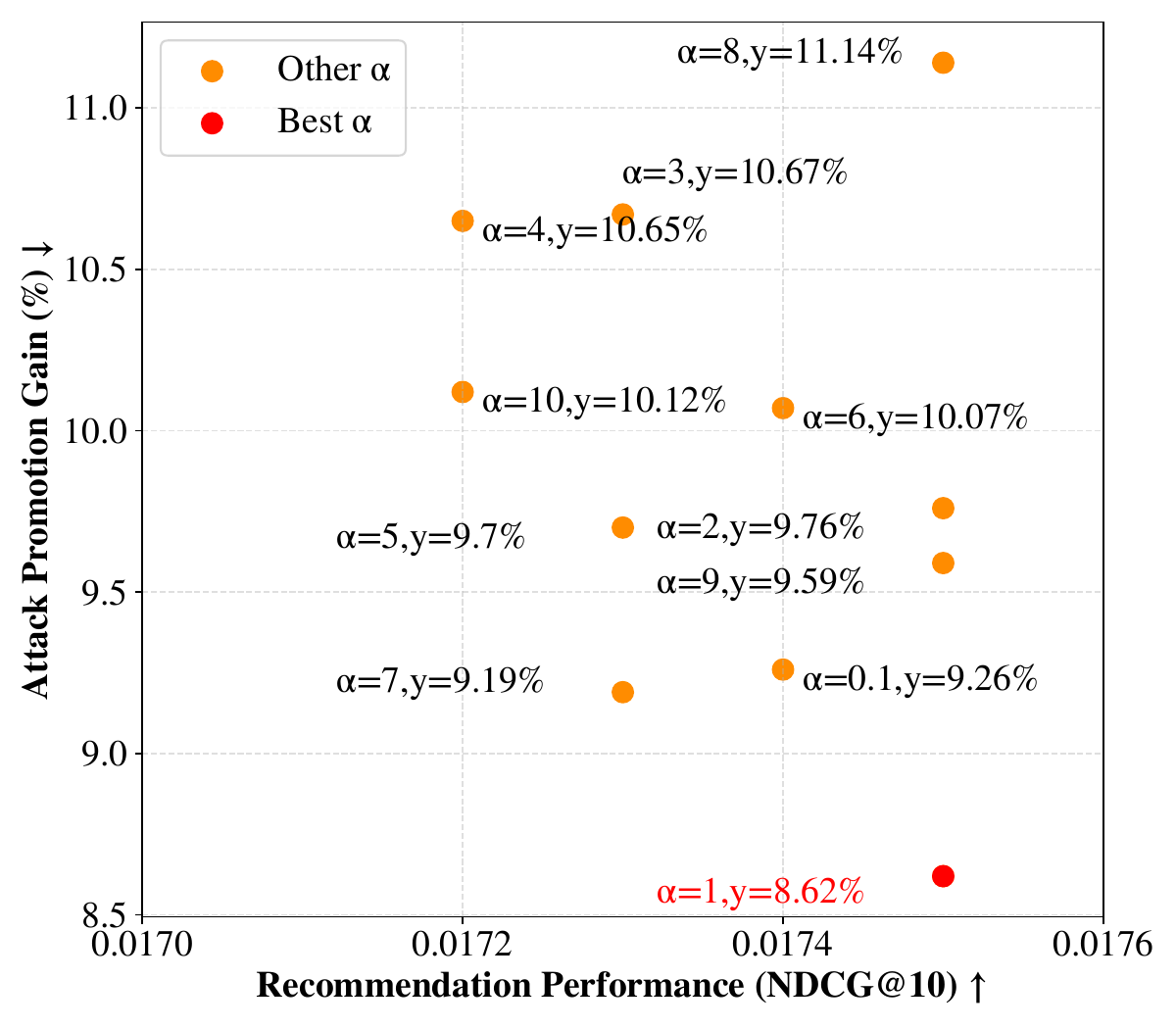}} \\
    
    \subfigure[MMGCN-Sports ($\lambda$)]{\includegraphics[width=0.45\columnwidth]{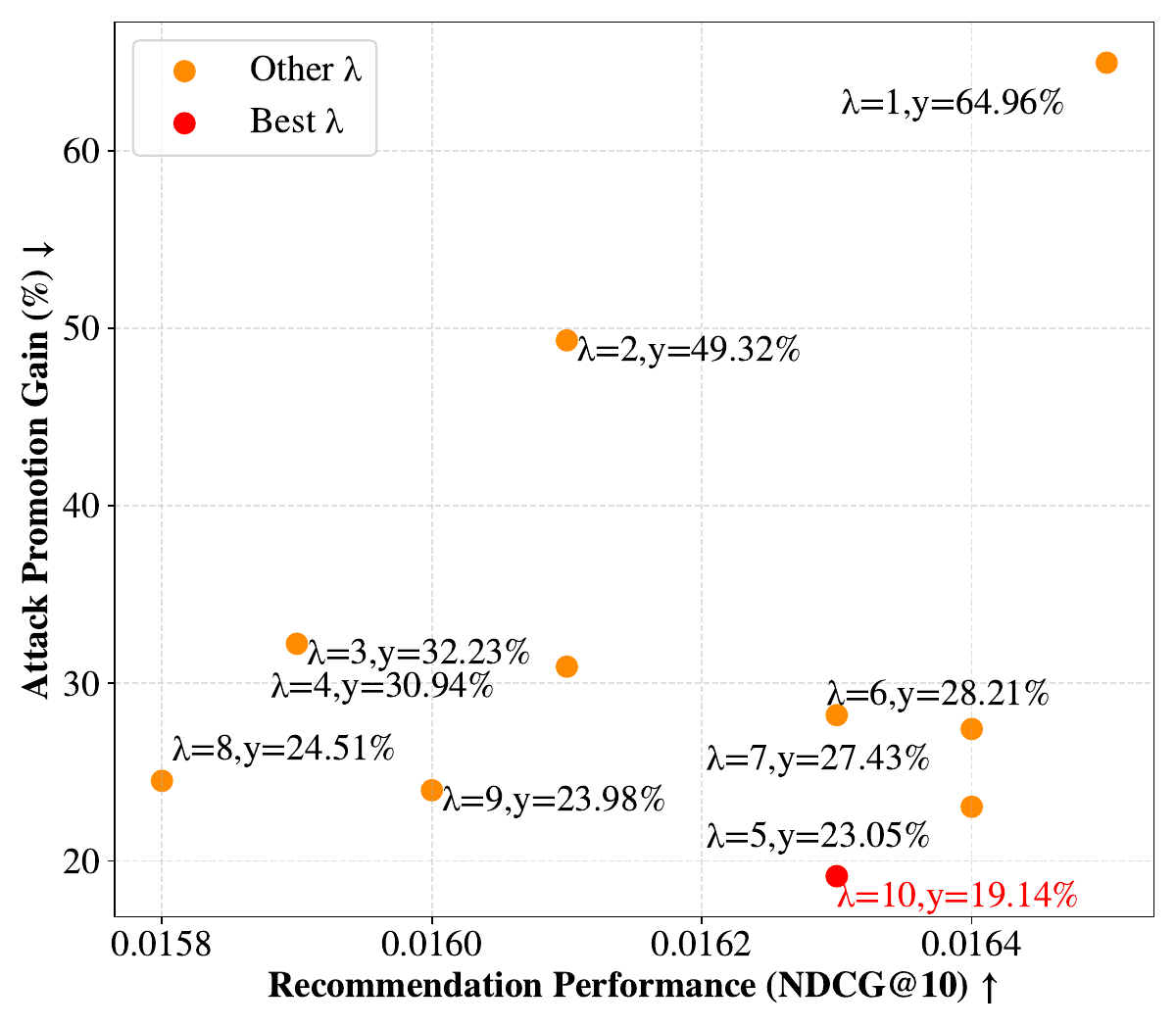}}\hfil
    \subfigure[MMGCN-Sports ($\alpha$)]{\includegraphics[width=0.45\columnwidth]{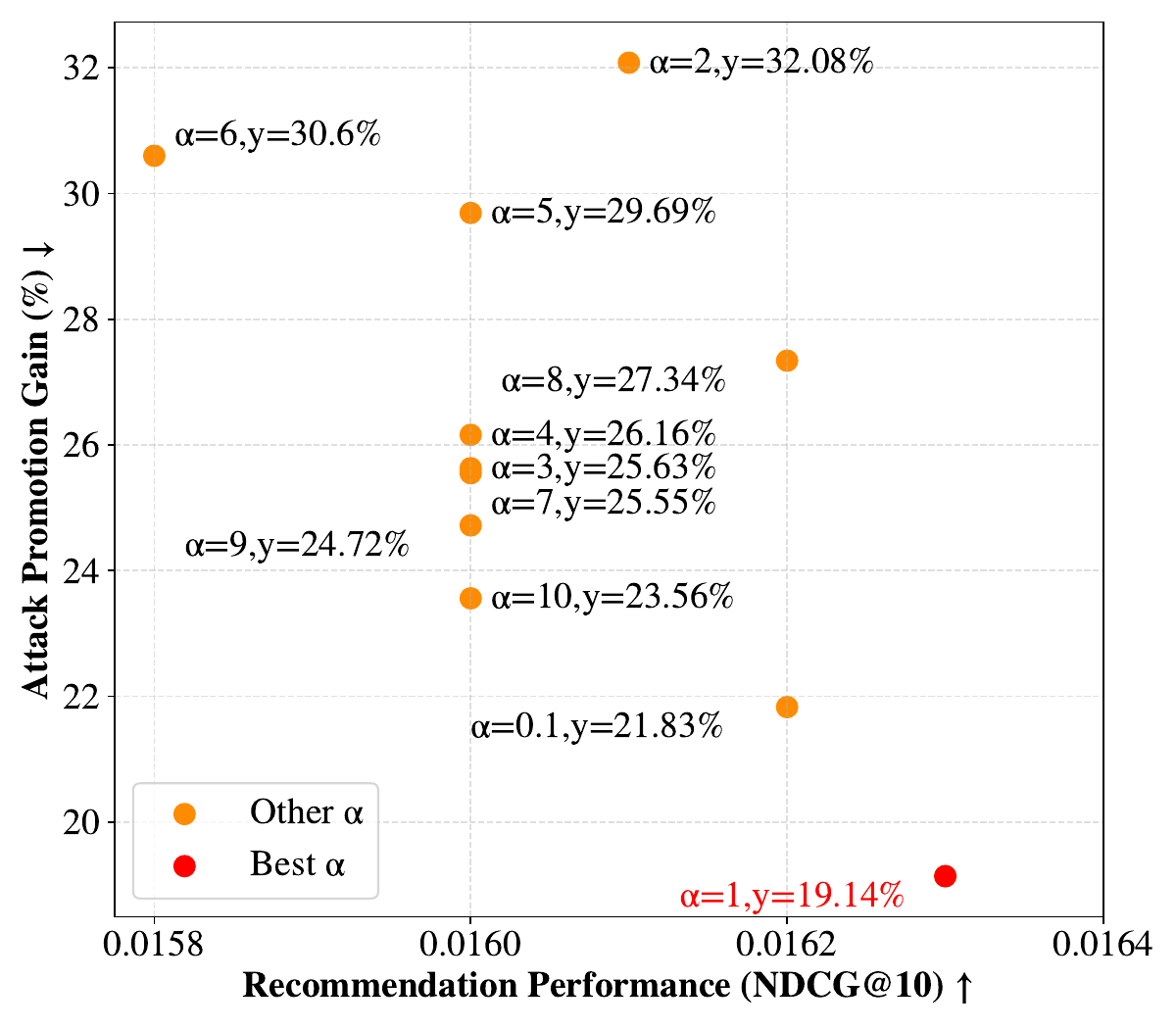}} \\
    
    \subfigure[MMGCN-Clothing ($\lambda$)]{\includegraphics[width=0.45\columnwidth]{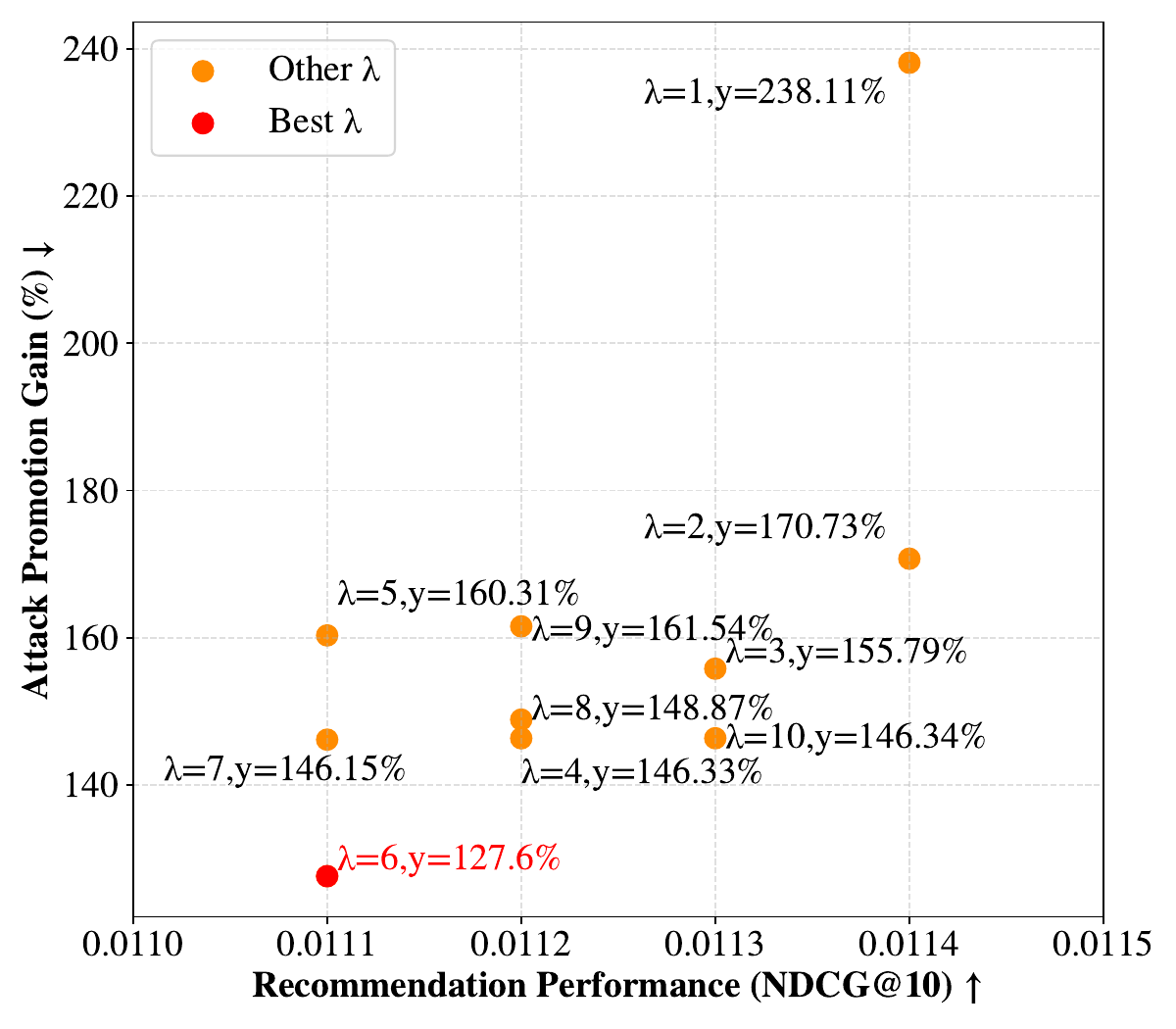}}\hfil
    \subfigure[MMGCN-Clothing ($\alpha$)]{\includegraphics[width=0.45\columnwidth]{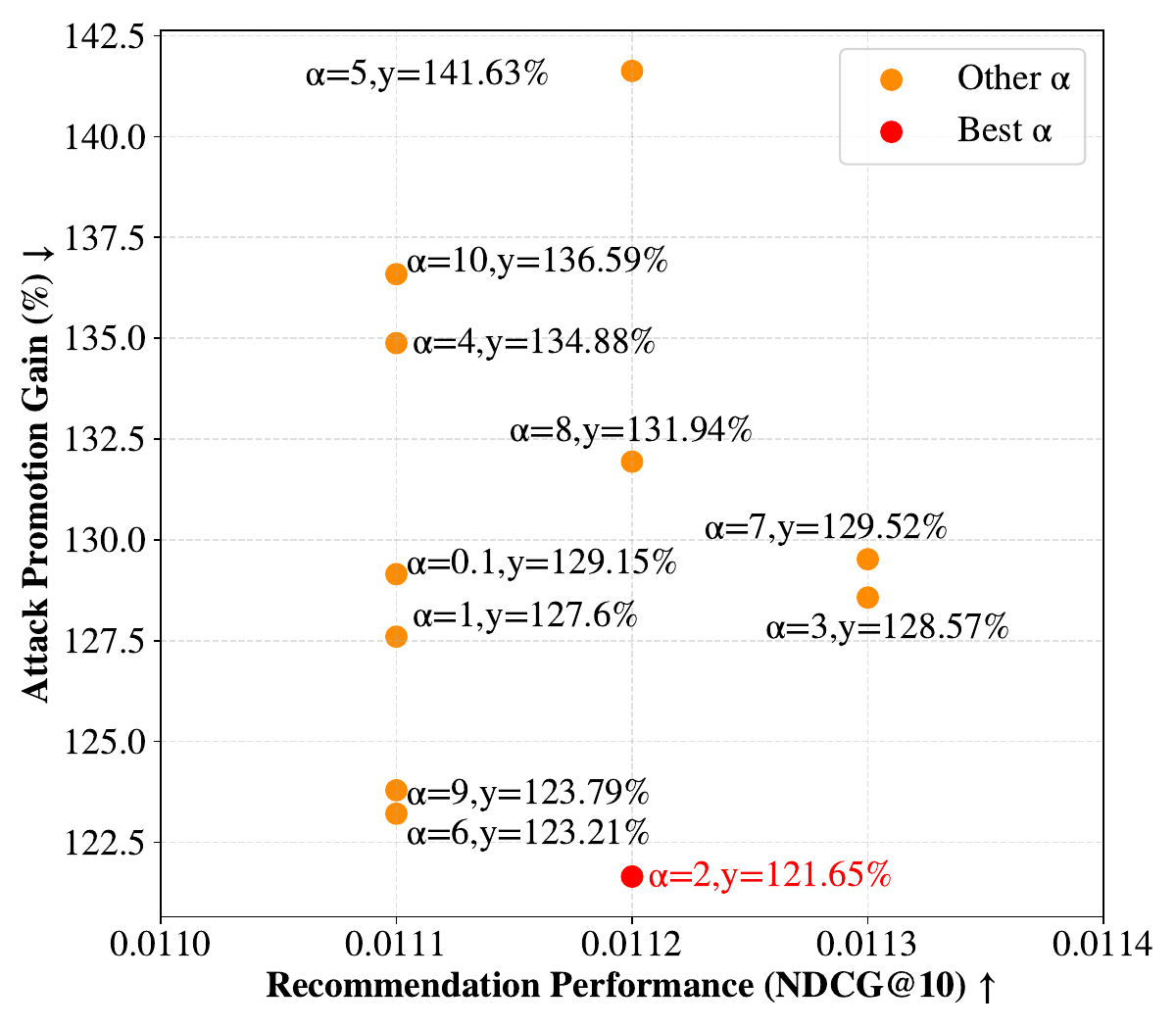}}
    
    \caption{Impact of $\lambda$ and $\alpha$  on the trade-off between accuracy and attack effectiveness for MMGCN under PGD-based attack. }
    \label{fig:MMGCN_PGD_Analysis}
\end{figure}

Figures~\ref{fig:VBPR_PGD_Analysis} and~\ref{fig:MMGCN_PGD_Analysis} illustrate the impact of hyperparameters on the trade-off between recommendation performance and adversarial robustness under PGD-based promotion attacks for VBPR and MMGCN, respectively. Specifically, the left columns of both figures demonstrate the effect of $\lambda$, while the right columns present the influence of $\alpha$, which controls the alignment loss between visual and textual gradients. The results are reported across three datasets: Amazon Baby (top row), Sports (middle row), and Clothing (bottom row).

\subsection{Extended Analysis of User Group Overlap}
\label{Appendix:jaccard_dist}

To provide a comprehensive empirical basis for the \textit{cross-modal gradient mismatch} identified in Section 3, we present the complete distribution of Jaccard Similarity coefficients between the visually-sensitive user subset $\mathcal{U}_v$ and the textually-sensitive user subset $\mathcal{U}_t$ across all three datasets (Amazon Baby, Sports, and Clothing) and two victim models (VBPR and MMGCN).

\begin{figure}[htbp]
	\centering
	\subfigure[VBPR-Baby]{\includegraphics[width=0.45\columnwidth]{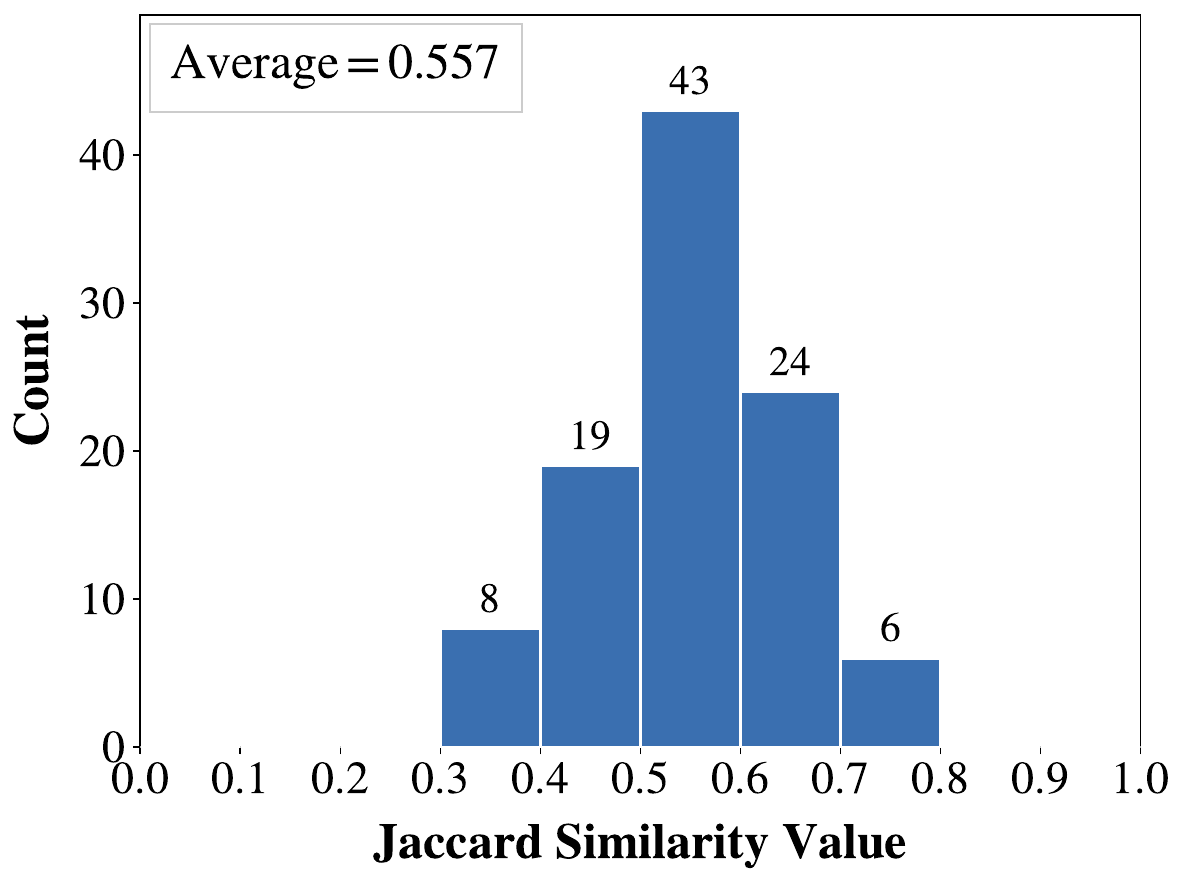}}\hfil
	\subfigure[MMGCN-Baby]{\includegraphics[width=0.45\columnwidth]{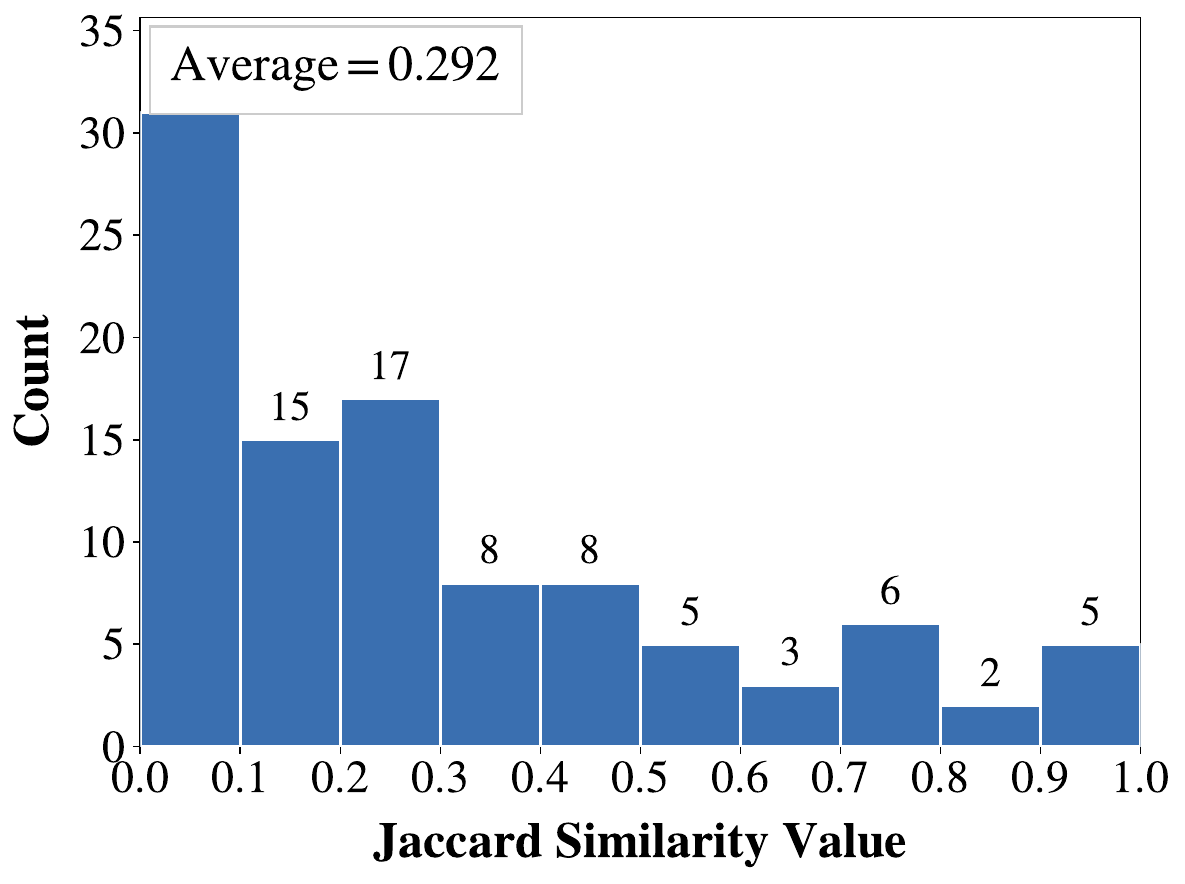}} \\
    \subfigure[VBPR-Sports]{\includegraphics[width=0.45\columnwidth]{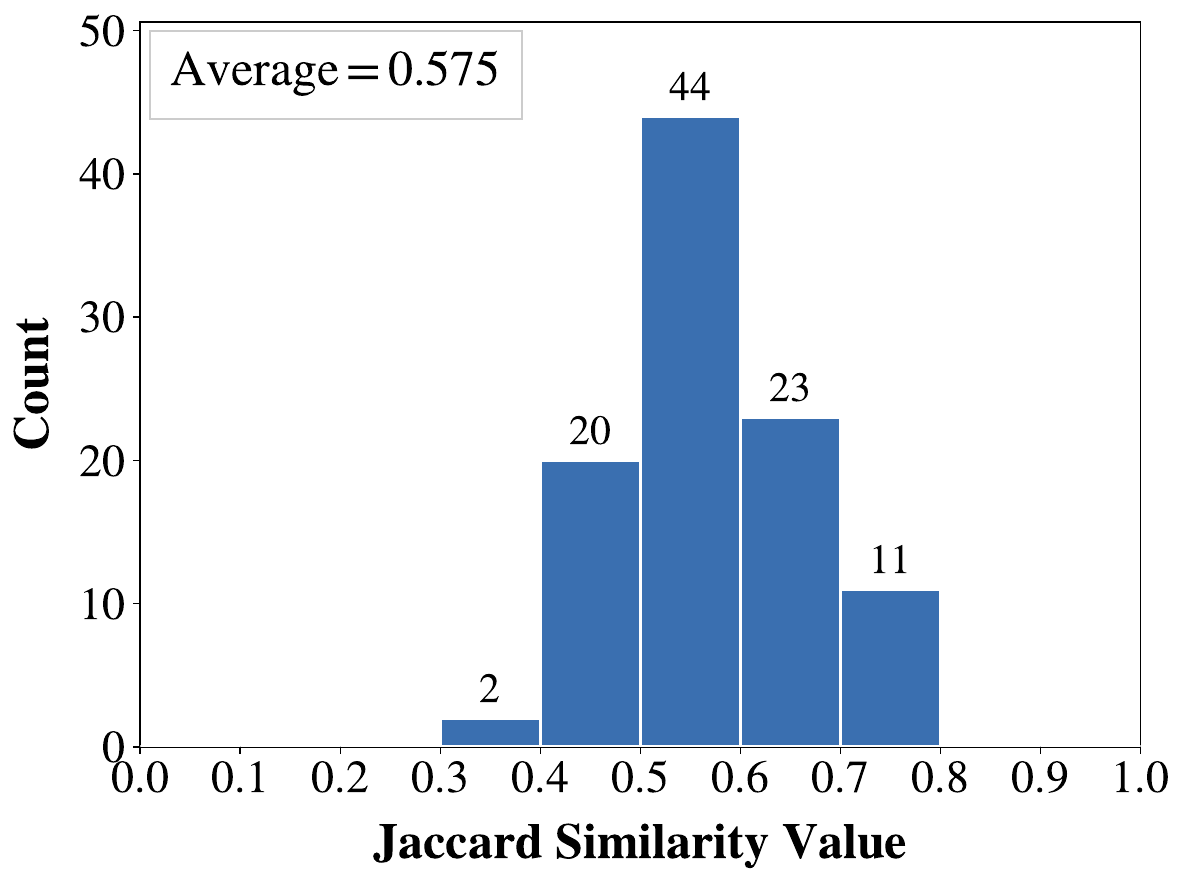}}\hfil
	\subfigure[MMGCN-Sports]{\includegraphics[width=0.45\columnwidth]{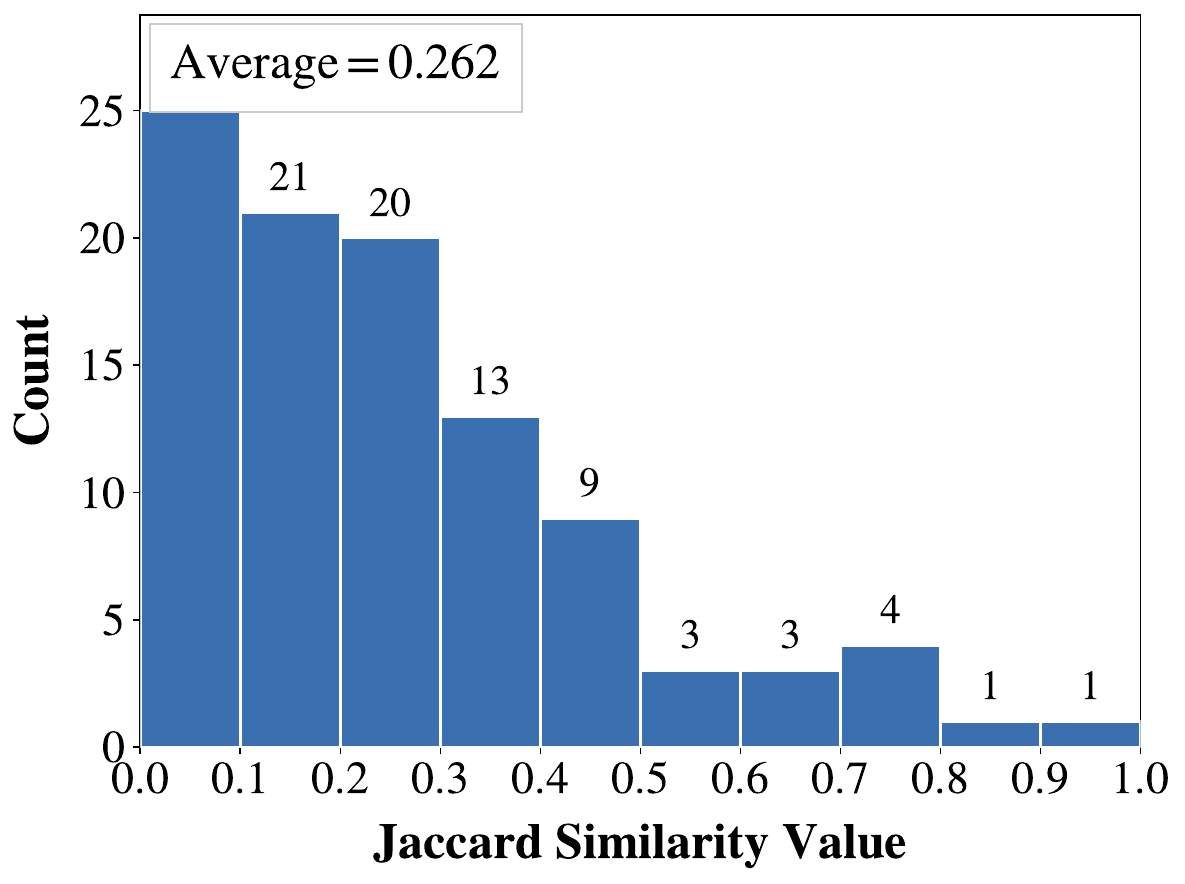}} \\
    \subfigure[VBPR-Clothing]{\includegraphics[width=0.45\columnwidth]{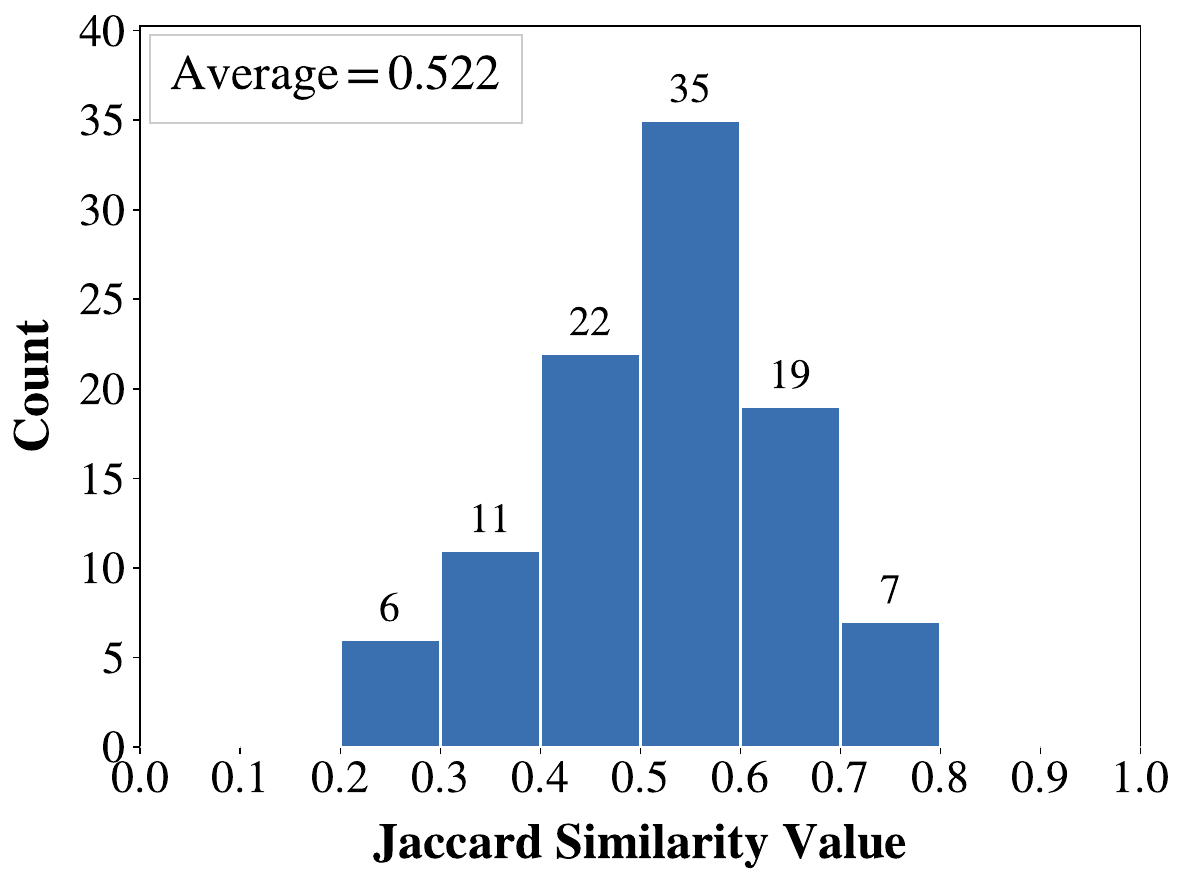}}\hfil
	\subfigure[MMGCN-Clothing]{\includegraphics[width=0.45\columnwidth]{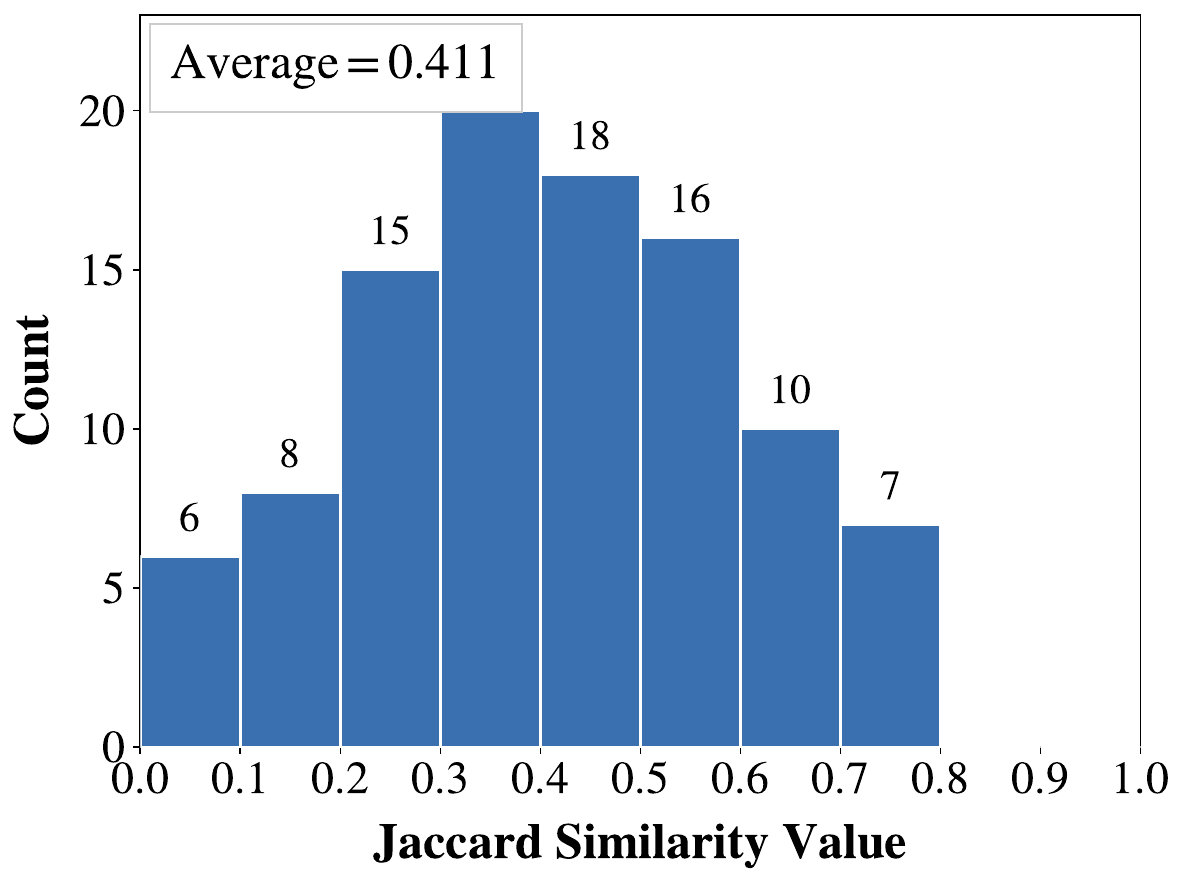}} 
	\caption{Distribution of Jaccard Similarity between $\mathcal{U}_v$ and $\mathcal{U}_t$.}
	\label{Fig-all-jaccard_distribution}
\end{figure}

Figure \ref{Fig-all-jaccard_distribution} illustrates these distributions.  Across all six experimental settings, the average Jaccard similarity remains consistently low to moderate, ranging from $0.262$ to $0.575$. This indicates that $\mathcal{U}_v$ and $\mathcal{U}_t$ are largely distinct groups. In other words, for a given targeted item, the users who are most susceptible to visual perturbations are rarely the same as those susceptible to textual perturbations. This distinctness fundamentally leads to the conflicting gradient directions during joint optimization. These extensive results further corroborate our motivation: without explicit coordination, multimodal perturbations naturally diverge due to the inherent discrepancy in user group dominance.

\subsection{Impact of $\epsilon_d$ and $\epsilon_a$}
\label{Appendix:appendix_budgets}

This section provides the detailed numerical results corresponding to the robustness analysis in the main paper. We evaluate the performance of UAT-MC under varying adversarial training budgets $\epsilon_d \in \{2.5\%, 5\%, 7.5\%, 10\%\}$ and attack budgets $\epsilon_a \in \{2.5\%, 5\%, 7.5\%, 10\%\}$.

The results are categorized based on the attack objective:
\begin{itemize}
    \item \textbf{Table~\ref{table:diff_epsilons}} presents the results under standard FGSM-based promotion attacks without the alignment loss.
    \item \textbf{Table~\ref{table:diff_epsilons_fgsm_align}} reports the $Gain_{Hit@50}$ under FGSM-based promotion attacks incorporating the gradient alignment loss $\mathcal{L}_{\text{Align}}$. These values correspond to the visualization in Figure~\ref{fig:diff_epsilon_fgsm_align}.
    
\end{itemize}
Across both settings, we observe that increasing the defense budget $\epsilon_d$ consistently suppresses the attack gain, demonstrating the robustness of our method against different variations of promotion attacks.

\subsection{Case Study}
\label{Appendix:case_study}
To illustrate how cross-modal gradient mismatch manifests during optimization and how gradient-level alignment reshapes the optimization dynamics,
we randomly select an item $i$ (ID: \texttt{B004203QQ4}, with 5 interactions) from the Baby dataset and conduct promotion attacks on VBPR.
We compare PGD w/o $\mathcal{L}_{\text{Align}}$ and PGD w/ $\mathcal{L}_{\text{Align}}$, tracking (i) the user coverage, measured by $N^i_{\text{rec}}$, and (ii) the cosine similarity between visual and textual perturbation gradients.

\begin{figure*}[t] 
  \centering
  \subfigure[User coverage of item $i$ ($N^i_{\text{rec}}$)]{
    \includegraphics[width=0.48\linewidth]{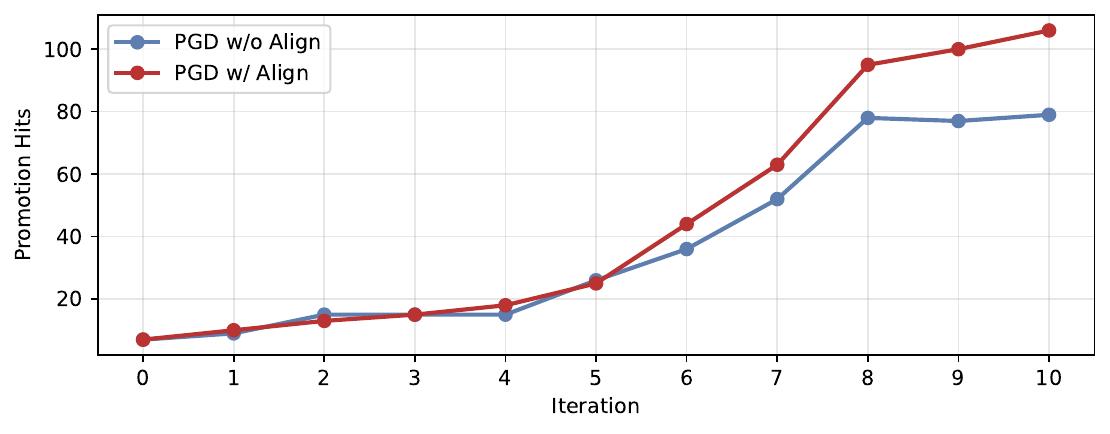}
  }
  \hfill
  \subfigure[Cosine similarity between $\Gamma_v$ and $\Gamma_t$.]{
    \includegraphics[width=0.48\linewidth]{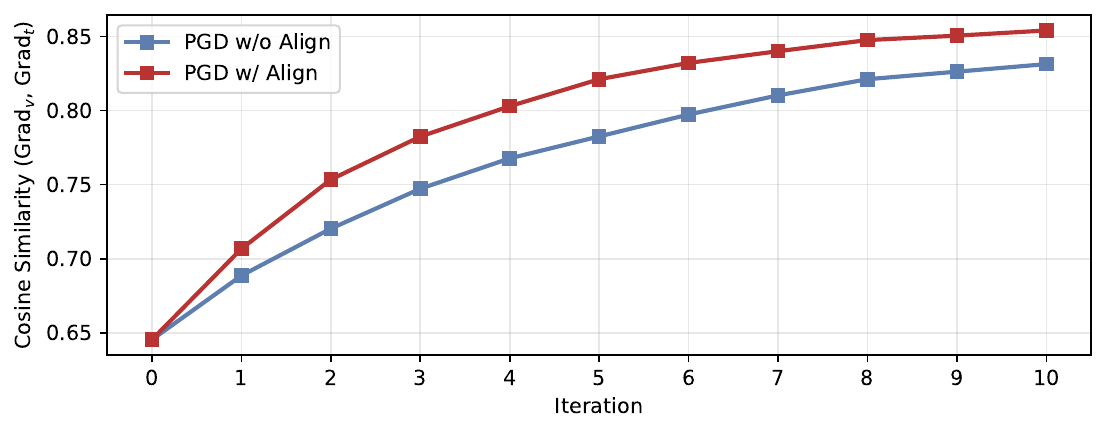}
  }

  \caption{Case study of PGD-based promotion attacks on item \texttt{B004203QQ4} from the Baby dataset using VBPR, with and without the alignment loss $\mathcal{L}_{\text{Align}}$.}
  \label{fig:case_study}
\end{figure*}

As shown in Fig.~\ref{fig:case_study}, vanilla PGD gradually increases the cross-modal gradient cosine similarity but quickly saturates at a relatively low level, resulting in slower and less stable growth in promotion performance. In contrast, when $\mathcal{L}_{\text{Align}}$ is introduced, the gradient cosine similarity rises more rapidly and remains consistently higher. Consequently, PGD with $\mathcal{L}_{\text{Align}}$ achieves faster and more stable promotion gains. Overall, this case study provides intuitive evidence that explicitly enforcing gradient alignment mitigates cross-modal gradient mismatch and leads to more effective promotion attacks.

\input{Tables/table_diff_epsilon}

\input{Tables/table_diff_epsilon_fgsm_align}

\section{Computational Complexity Analysis}
 Let $T_f$ and $T_b$ denote the forward and backward cost of the underlying MRS for one batch, $d$ the embedding dimension, and $K$ the nu mber of inner maximization steps. Without adversarial training, the per-batch complexity is $O(T_f+T_b)$.  
If one explicitly computes per-user gradients for both modalities to identify $U_v$ and $U_t$, the cost becomes $O(|U_p|T_b + |U_p|d + |U_p|\log |U_p|)\approx O(|U_p|T_b)$ per attack step, i.e., $O(K|U_p|T_b + T_f + T_b)$ per batch, which grows linearly with the number of users. In contrast, UAT-MC does not explicitly identify $U_v$ or $U_t$. It only adds a gradient-alignment term on top of the already available batch-level modality gradients, introducing an extra cost of only $O(d)$. Therefore, its per-batch complexity is approximately $O((K+1)(T_f+T_b))$, i.e., the same order as standard adversarial training without explicit user-wise coordination. Hence, compared with explicit user-wise gradient analysis, UAT-MC removes the linear dependence on $|U_p|$ and introduces only a lightweight extra overhead.

\end{document}

%% file: Tables/table_defense_result.tex
\begin{table*}[!t]

	\begin{center}
		\renewcommand{\arraystretch}{1}
		\tabcolsep=0.12cm
		\resizebox{\textwidth}{!}
		{
			\begin{tabular}{|c|c|c|c|cc|cc|cc|cc|cc|}
				\hline
				\multirow{3}{*}{Dataset} & \multirow{3}{*}{\shortstack{Victim\\Model}} & \multirow{3}{*}{\shortstack{Defense\\Method}} \
				& \multirow{3}{*}{$\text{Hit}_{\text{before}}$} \
				& \multicolumn{4}{c|}{FGSM-based Attack}  
				& \multicolumn{4}{c|}{PGD-based Attack} 
				& \multicolumn{2}{c|}{Recommendation} \\
				\cline{5-14} 
				&    &   &  
				         &  \multicolumn{2}{c|}{$\mathcal{L}_{\text{prom}}$}
				         &  \multicolumn{2}{c|}{$\mathcal{L}_{\text{prom}}+\mathcal{L}_{\text{Align}}$}
				          &  \multicolumn{2}{c|}{$\mathcal{L}_{\text{prom}}$}   
				         &  \multicolumn{2}{c|}{$\mathcal{L}_{\text{prom}}+\mathcal{L}_{\text{Align}}$}  
				         & \multirow{2}{*}{Recall} 
				         &\multirow{2}{*}{NDCG} 
				         \\
				   \cline{5-12} 
				  &    &   &  
				         & \shortstack{$\text{Hit}_{\text{after}}$}   
				         & \shortstack{$\text{Gain}_{\text{Hit}}$}    
				         & \shortstack{$\text{Hit}_{\text{after}}$}   
				         & \shortstack{$\text{Gain}_{\text{Hit}}$}    
				         & \shortstack{$\text{Hit}_{\text{after}}$}   
				         & \shortstack{$\text{Gain}_{\text{Hit}}$}  
				         & \shortstack{$\text{Hit}_{\text{after}}$}   
				         & \shortstack{$\text{Gain}_{\text{Hit}}$}   
				         &  
				         & 
				         \\
				         
				\hline
				\multirow{6}{*}{Baby}
				& \multirow{3}{*}{VBPR}
				  & w/o AT   & 0.667\% & 3.865\% & 479.38\% & 3.908\% & 485.80\% & 4.111\% & 516.15\% & 4.143\% & 520.99\% & 0.0503 & 0.027 \\
				& & UAT        & 0.622\% & 2.381\% & 282.96\% & 2.383\% & 283.41\% & 2.450\% & 294.15\% & 2.451\% & 294.25\% & 0.0484 & 0.0264 \\
				& & UAT-MC     & 0.621\% & 1.515\% & \textbf{143.98}\% & 1.515\% &  \textbf{144.00}\% & 1.544\% &  \textbf{148.65}\% & 1.544\% &  \textbf{148.65}\% & 0.0476 & 0.0253 \\
				\cline{2-14} 
				& \multirow{3}{*}{MMGCN}
				  & w/o AT   & 0.622\% & 1.747\% & 181.03\% & 1.748\% & 181.14\% & 3.399\% & 446.76\% & 3.421\% & 450.30\% & 0.0413 & 0.022 \\
				& & UAT        & 0.472\% & 0.507\% & 7.29\% & 0.508\% & 7.49\% & 0.515\% & 9.12\% & 0.516\% & 9.34\% & 0.0344 & 0.0185 \\
				& & UAT-MC     & 0.479\% & 0.513\% &  \textbf{7.21}\% & 0.514\% &  \textbf{7.47}\% & 0.519\% &  \textbf{8.46}\% & 0.520\% &  \textbf{8.62}\% & 0.0341 & 0.0185 \\
				
				\hline
				
				\multirow{6}{*}{Sports}
				& \multirow{3}{*}{VBPR}
				  & w/o AT   & 0.025\% & 6.547\% & 25722.99\% & 6.586\% & 25876.73\% & 7.253\% & 28506.93\% & 7.279\% & 28610.25\% & 0.0586 & 0.0319 \\
				& & UAT        & 0.025\% & 0.028\% & 13.45\% & 0.028\% & 13.45\% & 0.029\% & 13.73\% & 0.029\% & 13.73\% & 0.0510 & 0.0280 \\
				& & UAT-MC     & 0.024\% & 0.026\% &  \textbf{10.32}\% & 0.026\% &  \textbf{10.32}\% & 0.026\% &  \textbf{10.32}\% & 0.026\% &  \textbf{10.32}\% & 0.0508 & 0.0280 \\
				\cline{2-14} 
				& \multirow{3}{*}{MMGCN}
			   	  & w/o AT   & 0.024\% & 0.266\% & 1024.04\% & 0.268\% & 1033.23\% & 0.572\% & 2315.13\% & 0.581\% & 2352.82\% & 0.0385 & 0.0206 \\
				& & UAT        & 0.035\% & 0.038\% & 7.01\% & 0.039\% & 11.02\% & 0.044\% & 24.45\% & 0.044\% & 25.45\% & 0.0313 & 0.0172 \\
				& & UAT-MC     & 0.034\% & 0.034\% &  \textbf{0.00}\% & 0.037\% &  \textbf{8.44}\% & 0.041\% &  \textbf{19.14}\% & 0.041\% &  \textbf{19.14}\% & 0.0317 & 0.0173 \\
				
				\hline
				
				\multirow{6}{*}{Clothing}
				& \multirow{3}{*}{VBPR}
				  & w/o AT   & 0.036\% & 1.335\% & 3558.26\% & 1.380\% & 3682.09\% & 1.419\% & 3788.87\% & 1.438\% & 3838.61\% & 0.0384 & 0.0211 \\
				& & UAT        & 0.031\% & 0.287\% & 818.66\% & 0.303\% & 869.17\% & 0.306\% & 878.50\% & 0.312\% & 898.38\% & 0.0343 & 0.0189 \\
				& & UAT-MC     & 0.030\% & 0.038\% &  \textbf{26.64}\% & 0.038\% &  \textbf{26.85}\% & 0.038\% &  \textbf{27.70}\% & 0.038\% &  \textbf{27.70}\% & 0.0333 & 0.0184 \\
				\cline{2-14} 
				& \multirow{3}{*}{MMGCN}
				  & w/o AT   & 0.014\% & 1.886\% & 13282.43\% & 1.941\% & 13670.72\% & 9.035\% & 64019.37\% & 9.134\% & 64715.32\% & 0.0239 & 0.0123 \\
				& & UAT        & 0.014\% & 0.025\% & 76.00\% & 0.026\% & 81.78\% & 0.035\% & 148.44\% & 0.036\% & 149.33\% & 0.0221 & 0.0114 \\
				& & UAT-MC     & 0.015\% & 0.019\% &  \textbf{29.44}\% & 0.020\% &  \textbf{33.77}\% & 0.032\% &  \textbf{119.05}\% & 0.032\% &  \textbf{121.65}\% & 0.0221 & 0.0115 \\
			
				\hline
				
			\end{tabular}
		}

	\end{center}
		\caption{
		Performance of promotion attacks on two classical models. 
		All reported numbers are averaged results. The best defense results are highlighted in bold.
        (Note: $\text{Gain}_{\text{Hit}@50}$ values are computed using unrounded $\text{Hit}@50$ scores).
	}
	\label{table:defense_result}
	\vspace{-3mm}
\end{table*}

%% file: Tables/tables_rs_hyperparameter_search_space.tex
		
		
\begin{table}[t]
    \centering
    \small 
    \renewcommand{\arraystretch}{1.2} 
    \setlength{\tabcolsep}{4pt}
    
    \begin{tabular}{l p{0.65\columnwidth}}
        \toprule
        \textbf{Hyperparameter} & \textbf{Value / Search Space} \\
        \midrule
        \multicolumn{2}{c}{\textit{Common Settings (Shared)}} \\
        \midrule
        Embedding Size & $64$ \\
        Epochs & $[1, 1000]$ \\
        Stopping Step & $100$ \\
        Batch Size & $2048$ (Train), $4096$ (Eval) \\
        Optimizer & Adam \\
        LR Scheduler & $[1.0, 50]$ \\
        Align Weight & $\{2.0, 1.0, 0.1, 0.01, 0.001, 0\}$ \\
        \midrule
        \multicolumn{2}{c}{\textit{VBPR Specific}} \\
        \midrule
        Learning Rate & $0.001$ \\
        $\beta$ & $\{2.0, 1.0, 0.1, 0.01, 0.001, 10^{-4}, 10^{-5}\}$ \\
        \midrule
        \multicolumn{2}{c}{\textit{MMGCN Specific}} \\
        \midrule
        Layers & $2$ \\
        Learning Rate & $\{0.0001, 0.0005, 0.001, 0.005, 0.01\}$ \\
        $\beta$ & $\{10^{-5}, 10^{-4}, 0.001, 0.01, 0.1, 0.5\}$ \\
        \bottomrule
    \end{tabular}
    \caption{Hyper-parameter search spaces of multimodal recommendation systems.}
    \label{Table:hyper_parameter}
\end{table}

%% file: Tables/table_diff_epsilon.tex
\begin{table*}[!t]
	\begin{center}
		\small
		\renewcommand{\arraystretch}{1}
		\tabcolsep=0.12cm
		\footnotesize
		\resizebox{\textwidth}{!}
		{
			\begin{tabular}{|c|c|cccc|cccc|}
				\hline
				
				\multirow{2}{*}{Dataset} & Victim Model & \multicolumn{4}{c|}{VBPR} & \multicolumn{4}{c|}{MMGCN} \\
				\cline{2-10}

				& {$\epsilon$}   & {\rule{0pt}{2ex}$\epsilon_{a}$}=2.5\%   & {$\epsilon_{a}$}=5\%   & {$\epsilon_{a}$}=7.5\%    & {$\epsilon_{a}$}=10\%   & {$\epsilon_{a}$}=2.5\%   & {$\epsilon_{a}$}=5\%   & {$\epsilon_{a}$}=7.5\%    & {$\epsilon_{a}$}=10\%  \\
				\hline
				\multirow{5}{*}{Baby}
				& w/o AT & 54.25\% & 146.19\% & 288.11\% & 479.38\% & 81.49\% & 152.20\% & 176.08\% & 181.03\% \\
				& {$\epsilon_{d}$}=2.5\%    & 49.33\% & 138.78\% & 270.46\% & 456.33\% & 12.39\% & 21.72\% & 27.69\% & 31.13\% \\
				& {$\epsilon_{d}$}=5\%    & 43.01\% & 122.32\% & 243.43\% & 412.88\% & 5.65\% & 9.30\% & 13.07\% & 15.82\% \\
				& {$\epsilon_{d}$}=7.5\%    & 28.82\% & 73.92\% & 142.97\% & 233.41\% & 3.97\% & 7.37\% & 10.82\% & 12.94\% \\
				& {$\epsilon_{d}$}=10\%      & \textbf{19.38}\% & \textbf{48.45}\% &\textbf{89.52}\% & \textbf{143.98}\% & \textbf{2.42}\% & \textbf{4.32}\% & \textbf{5.83}\% & \textbf{7.21}\% \\
				\cline{1-10} 
				\multirow{5}{*}{Sports}
				& w/o AT & 734.90\% & 4636.57\% & 13506.93\% & 25722.99\% & 201.19\% & 505.93\% & 789.32\% & 1024.04\% \\
				& {$\epsilon_{d}$}=2.5\%    & 277.40\% & 1946.12\% & 6601.83\% & 14000.23\% & 30.33\% & 48.74\% & 51.46\% & 45.19\% \\
				& {$\epsilon_{d}$}=5\%    & 58.84\% & 292.82\% & 954.70\% & 2356.63\% & 16.90\% & 21.90\% & 26.43\% & 27.14\% \\
				& {$\epsilon_{d}$}=7.5\%    & 6.90\% & 14.37\% & 22.13\% & 34.48\% & 7.32\% & 8.87\% & 9.98\% & 10.42\% \\
				& {$\epsilon_{d}$}=10\%      & \textbf{2.36}\% & \textbf{4.72}\% & \textbf{7.37}\% & \textbf{10.32}\% & \textbf{0.00}\% & \textbf{3.91}\% & \textbf{4.53}\% & \textbf{6.38}\% \\
				\cline{1-10} 
				\multirow{5}{*}{Clothing}
				& w/o AT & 121.04\% & 519.48\% & 1508.35\% & 3558.26\% & 1072.07\% & 6460.36\% & 12045.05\% & 13282.43\% \\
				& {$\epsilon_{d}$}=2.5\%    & 79.76\% & 327.66\% & 1004.01\% & 2486.97\% & 128.08\% & 252.71\% & 303.45\% & 305.42\% \\
				& {$\epsilon_{d}$}=5\%    & 33.95\% & 100.20\% & 244.99\% & 522.09\% & 43.98\% & 61.28\% & 68.42\% & 71.80\% \\
				& {$\epsilon_{d}$}=7.5\%    & 13.35\% & 36.23\% & 67.37\% & 123.52\% & 30.04\% & 48.50\% & 51.93\% & 56.22\% \\
				& {$\epsilon_{d}$}=10\%      & \textbf{2.96}\% & \textbf{9.30}\% & \textbf{16.70}\% & \textbf{26.64}\% & \textbf{21.65}\% & \textbf{25.06}\% & \textbf{28.04}\% & \textbf{29.44}\% \\
				\cline{1-10} 
				\hline
				
			\end{tabular}
		}
	\end{center}
	\caption{
		Comparison of $\text{Gain}_{\text{Hit}@50}$ under different combinations of adversarial training budgets ($\epsilon_d$) and attack budgets ($\epsilon_a$) for FGSM-based promotion attacks \emph{without} gradient alignment. Here, $\epsilon_d$ denotes the perturbation budget used during adversarial training, while $\epsilon_a$ represents the perturbation budget used during promotion attacks. The best defense performance under each setting is highlighted in bold.
	}
	\label{table:diff_epsilons}
\end{table*}

%% file: Tables/table_diff_epsilon_fgsm_align.tex
\begin{table*}[!t]
	\begin{center}
		\small
		\renewcommand{\arraystretch}{1}
		\tabcolsep=0.12cm
		\footnotesize
		\resizebox{\textwidth}{!}
		{
			\begin{tabular}{|c|c|cccc|cccc|}
				\hline
				
				\multirow{2}{*}{Dataset} & Victim Model & \multicolumn{4}{c|}{VBPR} & \multicolumn{4}{c|}{MMGCN} \\
				\cline{2-10}

				& {$\epsilon$}   & {\rule{0pt}{2ex}$\epsilon_{a}$}=2.5\%   & {$\epsilon_{a}$}=5\%   & {$\epsilon_{a}$}=7.5\%    & {$\epsilon_{a}$}=10\%   & {$\epsilon_{a}$}=2.5\%   & {$\epsilon_{a}$}=5\%   & {$\epsilon_{a}$}=7.5\%    & {$\epsilon_{a}$}=10\%  \\
				\hline
				\multirow{5}{*}{Baby}
				& w/o AT & 54.81\% & 148.29\% & 291.56\% & 485.80\% & 81.68\% & 152.74\% & 176.58\% & 181.14\% \\
				& {$\epsilon_{d}$}=2.5\%    & 49.33\% & 138.78\% & 270.46\% & 456.33\% & 12.55\% & 22.02\% & 28.63\% & 32.24\% \\
				& {$\epsilon_{d}$}=5\%    & 43.01\% & 122.32\% & 243.43\% & 412.92\% & 5.71\% & 9.42\% & 13.21\% & 15.97\% \\
				& {$\epsilon_{d}$}=7.5\%    & 28.82\% & 73.92\% & 142.97\% & 233.41\% & 3.97\% & 7.39\% & 10.85\% & 13.05\% \\
				& {$\epsilon_{d}$}=10\%      & \textbf{19.38}\% & \textbf{48.47}\% & \textbf{89.54}\% & \textbf{144.00}\% & \textbf{2.44}\% & \textbf{4.35}\% & \textbf{5.88}\% & \textbf{7.47}\% \\
				\cline{1-10} 
				\multirow{5}{*}{Sports}
				& w/o AT & 754.29\% \quad & 4737.40\% & 13737.12\% & 25876.73\% & 202.37\% & 512.76\% & 807.12\% & 1033.23\% \\
				& {$\epsilon_{d}$}=2.5\%    & 277.40\% & 1947.26\% & 6602.51\% & 14000.68\% & 31.59\% & 52.30\% & 52.72\% & 56.49\% \\
				& {$\epsilon_{d}$}=5\%    & 60.22\% & 296.96\% & 965.75\% & 2384.53\% & 17.38\% & 30.00\% & 30.48\% & 31.19\% \\
				& {$\epsilon_{d}$}=7.5\%    & 7.18\% & 14.66\% & 23.56\% & 35.34\% & 8.65\% & 11.75\% & 13.75\% & 14.63\% \\
				& {$\epsilon_{d}$}=10\%      & \textbf{2.36}\% & \textbf{4.72}\% & \textbf{7.37}\% & \textbf{10.32}\% & \textbf{4.94}\% & \textbf{6.58}\% & \textbf{7.82}\% & \textbf{8.44}\% \\
				\cline{1-10} 
				\multirow{5}{*}{Clothing}
				& w/o AT & 125.91\% & 534.78\% & 1554.78\% & 3682.09\% & 1091.44\% & 6598.65\% & 12292.34\% & 13670.72\% \\
				& {$\epsilon_{d}$}=2.5\%    & 79.76\% & 327.66\% & 1004.61\% & 2488.58\% & 132.02\% & 263.05\% & 327.59\% & 340.89\% \\
				& {$\epsilon_{d}$}=5\%    & 33.95\% & 100.61\% & 244.99\% & 522.70\% & 47.37\% & 70.30\% & 79.70\% & 80.83\% \\
				& {$\epsilon_{d}$}=7.5\%    & 13.35\% & 36.23\% & 67.58\% & 123.73\% & 32.19\% & 54.94\% & 57.08\% & 63.52\% \\
				& {$\epsilon_{d}$}=10\%      & \textbf{3.38}\% & \textbf{9.51}\% & \textbf{16.70}\% & \textbf{26.85}\% & \textbf{22.51}\% & \textbf{27.05}\% & \textbf{29.53}\% & \textbf{33.77}\% \\
				\cline{1-10} 
				\hline
				
			\end{tabular}
		}
	\end{center}
	\caption{
		Comparison of $\text{Gain}_{\text{Hit}@50}$ under different combinations of adversarial training budgets ($\epsilon_d$) and attack budgets ($\epsilon_a$) for FGSM-based promotion attacks \emph{with} the gradient alignment loss $\mathcal{L}_{\text{Align}}$. Here, $\epsilon_d$ denotes the perturbation budget used during adversarial training, while $\epsilon_a$ represents the perturbation budget used during promotion attacks. A larger value of $\text{Gain}_{\text{Hit}@50}$ indicates higher attack effectiveness and thus weaker defense capability. The best defense performance under each setting is highlighted in bold.
	}
	\label{table:diff_epsilons_fgsm_align}
\end{table*}

%% file: UAT_MC.bbl
\begin{thebibliography}{}

\bibitem[\protect\citeauthoryear{Chen \bgroup \em et al.\egroup }{2024}]{Chen_2024_Visually_IPDGI}
Lijian Chen, Wei Yuan, Tong Chen, Guanhua Ye, Nguyen Quoc~Viet Hung, and Hongzhi Yin.
\newblock Adversarial item promotion on visually-aware recommender systems by guided diffusion.
\newblock {\em ACM Trans. Inf. Syst.}, 42(6), August 2024.

\bibitem[\protect\citeauthoryear{Di~Noia \bgroup \em et al.\egroup }{2020}]{Di_2020_Targeted_TAaMR}
Tommaso Di~Noia, Daniele Malitesta, and Felice~Antonio Merra.
\newblock Taamr: Targeted adversarial attack against multimedia recommender systems.
\newblock In {\em 2020 50th Annual IEEE/IFIP International Conference on Dependable Systems and Networks Workshops (DSN-W)}, pages 1--8, 2020.

\bibitem[\protect\citeauthoryear{Goodfellow \bgroup \em et al.\egroup }{2014}]{goodfellow_2014_explaining_FGSM}
Ian~J Goodfellow, Jonathon Shlens, and Christian Szegedy.
\newblock Explaining and harnessing adversarial examples.
\newblock {\em arXiv preprint arXiv:1412.6572}, 2014.

\bibitem[\protect\citeauthoryear{Guo \bgroup \em et al.\egroup }{2024}]{guo_2024_lgmrec_LGMRec}
Zhiqiang Guo, Jianjun Li, Guohui Li, Chaoyang Wang, Si~Shi, and Bin Ruan.
\newblock Lgmrec: local and global graph learning for multimodal recommendation.
\newblock In {\em Proceedings of the AAAI Conference on Artificial Intelligence}, volume~38, pages 8454--8462, 2024.

\bibitem[\protect\citeauthoryear{He and McAuley}{2016a}]{he_2016_ups_Amazon}
Ruining He and Julian McAuley.
\newblock Ups and downs: Modeling the visual evolution of fashion trends with one-class collaborative filtering.
\newblock In {\em proceedings of the 25th international conference on world wide web}, pages 507--517, 2016.

\bibitem[\protect\citeauthoryear{He and McAuley}{2016b}]{he_2016_VBPR}
Ruining He and Julian McAuley.
\newblock Vbpr: visual bayesian personalized ranking from implicit feedback.
\newblock In {\em Proceedings of the AAAI conference on artificial intelligence}, volume~30, 2016.

\bibitem[\protect\citeauthoryear{Hsiao \bgroup \em et al.\egroup }{2022}]{hsiao_2022_unsupervised_RecMR}
Shao-Ping Hsiao, Yu-Che Tsai, and Cheng-Te Li.
\newblock Unsupervised post-time fake social message detection with recommendation-aware representation learning.
\newblock In {\em Companion Proceedings of the Web Conference 2022}, pages 232--235, 2022.

\bibitem[\protect\citeauthoryear{Li \bgroup \em et al.\egroup }{2025}]{li_2025_Teach_GUIDER}
Hongji Li, Hanwen Du, Youhua Li, Junchen Fu, Chunxiao Li, Ziyi Zhuang, Jiakang Li, and Yongxin Ni.
\newblock Teach me how to denoise: A universal framework for denoising multi-modal recommender systems via guided calibration.
\newblock In {\em Proceedings of the Eighteenth ACM International Conference on Web Search and Data Mining}, WSDM '25, page 782–791, New York, NY, USA, 2025. Association for Computing Machinery.

\bibitem[\protect\citeauthoryear{Lin \bgroup \em et al.\egroup }{2023}]{lin2023autodenoise}
Weilin Lin, Xiangyu Zhao, Yejing Wang, Yuanshao Zhu, and Wanyu Wang.
\newblock Autodenoise: Automatic data instance denoising for recommendations.
\newblock In {\em Proceedings of the ACM Web Conference 2023}, pages 1003--1011, 2023.

\bibitem[\protect\citeauthoryear{Liu and Larson}{2021}]{Liu_2021_Adversarial_AIP_Image}
Zhuoran Liu and Martha Larson.
\newblock Adversarial item promotion: Vulnerabilities at the core of top-n recommenders that use images to address cold start.
\newblock In {\em Proceedings of the Web Conference 2021}, WWW '21, page 3590–3602, New York, NY, USA, 2021. Association for Computing Machinery.

\bibitem[\protect\citeauthoryear{Liu \bgroup \em et al.\egroup }{2024}]{Liu_2024_AlignRec}
Yifan Liu, Kangning Zhang, Xiangyuan Ren, Yanhua Huang, Jiarui Jin, Yingjie Qin, Ruilong Su, Ruiwen Xu, Yong Yu, and Weinan Zhang.
\newblock Alignrec: Aligning and training in multimodal recommendations.
\newblock In {\em Proceedings of the 33rd ACM International Conference on Information and Knowledge Management}, CIKM '24, page 1503–1512, New York, NY, USA, 2024. Association for Computing Machinery.

\bibitem[\protect\citeauthoryear{Madry \bgroup \em et al.\egroup }{2017}]{madry2017towards_PGD}
Aleksander Madry, Aleksandar Makelov, Ludwig Schmidt, Dimitris Tsipras, and Adrian Vladu.
\newblock Towards deep learning models resistant to adversarial attacks.
\newblock {\em arXiv preprint arXiv:1706.06083}, 2017.

\bibitem[\protect\citeauthoryear{Mu \bgroup \em et al.\egroup }{2025}]{mu_2025_trust_Trust-GRS}
Lingyu Mu, Zhengxiao Liu, Zhitong Zhu, and Zheng Lin.
\newblock Trust-grs: A trustworthy training framework for graph neural network based recommender systems against shilling attacks.
\newblock In {\em Proceedings of the AAAI Conference on Artificial Intelligence}, volume~39, pages 12408--12416, 2025.

\bibitem[\protect\citeauthoryear{Nguyen~Thanh \bgroup \em et al.\egroup }{2023}]{Thanh_2023_GSPAttack}
Toan Nguyen~Thanh, Nguyen Duc~Khang Quach, Thanh~Tam Nguyen, Thanh~Trung Huynh, Viet~Hung Vu, Phi~Le Nguyen, Jun Jo, and Quoc Viet~Hung Nguyen.
\newblock Poisoning gnn-based recommender systems with generative surrogate-based attacks.
\newblock {\em ACM Trans. Inf. Syst.}, 41(3), February 2023.

\bibitem[\protect\citeauthoryear{Ong and Khong}{2025}]{Ong_2025_Spectrum_SMORE}
Rongqing~Kenneth Ong and Andy W.~H. Khong.
\newblock Spectrum-based modality representation fusion graph convolutional network for multimodal recommendation.
\newblock In {\em Proceedings of the Eighteenth ACM International Conference on Web Search and Data Mining}, WSDM '25, page 773–781, New York, NY, USA, 2025. Association for Computing Machinery.

\bibitem[\protect\citeauthoryear{Rendle \bgroup \em et al.\egroup }{2009}]{Rendle_2009_BPR}
Steffen Rendle, Christoph Freudenthaler, Zeno Gantner, and Lars Schmidt-Thieme.
\newblock Bpr: Bayesian personalized ranking from implicit feedback.
\newblock In {\em Proceedings of the Twenty-Fifth Conference on Uncertainty in Artificial Intelligence}, UAI '09, page 452–461, Arlington, Virginia, USA, 2009. AUAI Press.

\bibitem[\protect\citeauthoryear{Sun \bgroup \em et al.\egroup }{2020}]{sun_2020_multi_MMKGs_MMGAT}
Rui Sun, Xuezhi Cao, Yan Zhao, Junchen Wan, Kun Zhou, Fuzheng Zhang, Zhongyuan Wang, and Kai Zheng.
\newblock Multi-modal knowledge graphs for recommender systems.
\newblock In {\em Proceedings of the 29th ACM international conference on information \& knowledge management}, pages 1405--1414, 2020.

\bibitem[\protect\citeauthoryear{Tang \bgroup \em et al.\egroup }{2020}]{Tang_2020_Adversarial_AMR}
Jinhui Tang, Xiaoyu Du, Xiangnan He, Fajie Yuan, Qi~Tian, and Tat-Seng Chua.
\newblock Adversarial training towards robust multimedia recommender system.
\newblock {\em IEEE Transactions on Knowledge and Data Engineering}, 32(5):855--867, 2020.

\bibitem[\protect\citeauthoryear{Wang \bgroup \em et al.\egroup }{2022}]{wang_2022_detecting_NFGCN-TIA}
Shilei Wang, Peng Zhang, Hui Wang, Hongtao Yu, and Fuzhi Zhang.
\newblock Detecting shilling groups in online recommender systems based on graph convolutional network.
\newblock {\em Information Processing \& Management}, 59(5):103031, 2022.

\bibitem[\protect\citeauthoryear{Wang \bgroup \em et al.\egroup }{2023}]{Wang_2023_DualGNN}
Qifan Wang, Yinwei Wei, Jianhua Yin, Jianlong Wu, Xuemeng Song, and Liqiang Nie.
\newblock Dualgnn: Dual graph neural network for multimedia recommendation.
\newblock {\em Multimedia, IEEE Trans. on (T-MM)}, 25(000):11, 2023.

\bibitem[\protect\citeauthoryear{Wang \bgroup \em et al.\egroup }{2024a}]{Wang_2024_llmpowered_TextSimu}
Zongwei Wang, Min Gao, Junliang Yu, Xinyi Gao, Quoc Viet~Hung Nguyen, Shazia Sadiq, and Hongzhi Yin.
\newblock Llm-powered text simulation attack against id-free recommender systems, 2024.

\bibitem[\protect\citeauthoryear{Wang \bgroup \em et al.\egroup }{2024b}]{wang_2024_unveiling_CLeaR}
Zongwei Wang, Junliang Yu, Min Gao, Hongzhi Yin, Bin Cui, and Shazia Sadiq.
\newblock Unveiling vulnerabilities of contrastive recommender systems to poisoning attacks.
\newblock In {\em Proceedings of the 30th ACM SIGKDD conference on knowledge discovery and data mining}, pages 3311--3322, 2024.

\bibitem[\protect\citeauthoryear{Wei \bgroup \em et al.\egroup }{2019}]{wei_2019_MMGCN}
Yinwei Wei, Xiang Wang, Liqiang Nie, Xiangnan He, Richang Hong, and Tat-Seng Chua.
\newblock Mmgcn: Multi-modal graph convolution network for personalized recommendation of micro-video.
\newblock In {\em Proceedings of the 27th ACM international conference on multimedia}, pages 1437--1445, 2019.

\bibitem[\protect\citeauthoryear{Wei \bgroup \em et al.\egroup }{2020}]{Wei_2020_Graph_GRCN}
Yinwei Wei, Xiang Wang, Liqiang Nie, Xiangnan He, and Tat-Seng Chua.
\newblock Graph-refined convolutional network for multimedia recommendation with implicit feedback.
\newblock In {\em Proceedings of the 28th ACM International Conference on Multimedia}, MM '20, page 3541–3549, New York, NY, USA, 2020. Association for Computing Machinery.

\bibitem[\protect\citeauthoryear{Wu \bgroup \em et al.\egroup }{2021}]{wu_2021_Fight_APT}
Chenwang Wu, Defu Lian, Yong Ge, Zhihao Zhu, Enhong Chen, and Senchao Yuan.
\newblock Fight fire with fire: Towards robust recommender systems via adversarial poisoning training.
\newblock In {\em Proceedings of the 44th International ACM SIGIR Conference on Research and Development in Information Retrieval}, SIGIR '21, page 1074–1083, New York, NY, USA, 2021. Association for Computing Machinery.

\bibitem[\protect\citeauthoryear{Wu \bgroup \em et al.\egroup }{2023}]{Wu_2023_Influence_Infmix}
Chenwang Wu, Defu Lian, Yong Ge, Zhihao Zhu, and Enhong Chen.
\newblock Influence-driven data poisoning for robust recommender systems.
\newblock {\em IEEE Trans. Pattern Anal. Mach. Intell.}, 45(10):11915–11931, October 2023.

\bibitem[\protect\citeauthoryear{Yang \bgroup \em et al.\egroup }{2024}]{Yang_2024_Attacking_SPAF}
Shiyi Yang, Chen Wang, Xiwei Xu, Liming Zhu, and Lina Yao.
\newblock Attacking visually-aware recommender systems with transferable and imperceptible adversarial styles.
\newblock In {\em Proceedings of the 33rd ACM International Conference on Information and Knowledge Management}, CIKM '24, page 2900–2909, New York, NY, USA, 2024. Association for Computing Machinery.

\bibitem[\protect\citeauthoryear{Yu \bgroup \em et al.\egroup }{2023}]{yu_2023_multi_MGCN}
Penghang Yu, Zhiyi Tan, Guanming Lu, and Bing-Kun Bao.
\newblock Multi-view graph convolutional network for multimedia recommendation.
\newblock In {\em Proceedings of the 31st ACM international conference on multimedia}, pages 6576--6585, 2023.

\bibitem[\protect\citeauthoryear{Zhang \bgroup \em et al.\egroup }{2021}]{zhang_2021_mining_LATTICE}
Jinghao Zhang, Yanqiao Zhu, Qiang Liu, Shu Wu, Shuhui Wang, and Liang Wang.
\newblock Mining latent structures for multimedia recommendation.
\newblock In {\em Proceedings of the 29th ACM international conference on multimedia}, pages 3872--3880, 2021.

\bibitem[\protect\citeauthoryear{Zhang \bgroup \em et al.\egroup }{2024}]{zhang_2024_stealthy_RecTextAttack}
Jinghao Zhang, Yuting Liu, Qiang Liu, Shu Wu, Guibing Guo, and Liang Wang.
\newblock Stealthy attack on large language model based recommendation.
\newblock In Lun-Wei Ku, Andre Martins, and Vivek Srikumar, editors, {\em Proceedings of the 62nd Annual Meeting of the Association for Computational Linguistics (Volume 1: Long Papers)}, pages 5839--5857, Bangkok, Thailand, August 2024. Association for Computational Linguistics.

\bibitem[\protect\citeauthoryear{Zhou \bgroup \em et al.\egroup }{2023a}]{Zhou_2023_A_MMRec}
Hongyu Zhou, Xin Zhou, Zhiwei Zeng, Lingzi Zhang, and Zhiqi Shen.
\newblock A comprehensive survey on multimodal recommender systems: Taxonomy, evaluation, and future directions, 2023.

\bibitem[\protect\citeauthoryear{Zhou \bgroup \em et al.\egroup }{2023b}]{zhou2023enhancing_DRAGON}
Hongyu Zhou, Xin Zhou, Lingzi Zhang, and Zhiqi Shen.
\newblock Enhancing dyadic relations with homogeneous graphs for multimodal recommendation.
\newblock In {\em ECAI 2023}, pages 3123--3130. IOS Press, 2023.

\bibitem[\protect\citeauthoryear{Zhou \bgroup \em et al.\egroup }{2023c}]{zhou_2023_bootstrap_BM3}
Xin Zhou, Hongyu Zhou, Yong Liu, Zhiwei Zeng, Chunyan Miao, Pengwei Wang, Yuan You, and Feijun Jiang.
\newblock Bootstrap latent representations for multi-modal recommendation.
\newblock In {\em Proceedings of the ACM web conference 2023}, pages 845--854, 2023.

\end{thebibliography}
